\documentclass[a4paper,fleqn]{cas-dc}

\usepackage[numbers]{natbib}
\usepackage{adjustbox}
\usepackage{tabularx}
\usepackage{float}
\usepackage{booktabs}
\usepackage{caption}
\usepackage{graphicx}
\usepackage{rotating}
\usepackage{subcaption}
\usepackage{longtable}
\usepackage{url}
\usepackage{lipsum}
\usepackage{geometry}
\usepackage{placeins}
\usepackage{afterpage}
\usepackage{hyperref}

\DeclareUnicodeCharacter{1F4AA}{}
\DeclareUnicodeCharacter{1FA7A}{\stethoscope}
\DeclareUnicodeCharacter{1FAC1}{\lungs}

\begin{document}

%\let\WriteBookmarks\relax
%\def\floatpagepagefraction{1}
%\def\textpagefraction{.001}

\shorttitle{A.T Otapo et al. Prediction and Detection of Terminal Diseases Using Using Internet of Medical Things: A Review}

\shortauthors{A.T Otapo et al.}

\title[mode=title]{Prediction and Detection of Terminal Diseases Using Internet of Medical Things: A Review}

% Title footnote texts
\thanks{This document is the result of the research project carried out at Laboratoire Images, Signaux et Systémes Intelligents (LiSSi)–EA 3956, Université Paris-Est Créteil (UPEC), 94010 Créteil Cedex, France.}

\author[a]{Akeem Temitope Otapo}[type=editor,
                        auid=000,bioid=1,
                        role=]
\ead{akeem.otapo@u-pec.fr}
\ead[url]{https://orcid.org/0000-0002-3032-0393}
\credit{Conceptualization, Methodology, Investigation, Formal Analysis, Writing - Original Draft}

\author[a]{Alice Othmani}[style=chinese]
%\cormark[1]

\ead{alice.othmani@u-pec.fr}
\ead[url]{https://orcid.org/0000-0001-5898-1196}
\credit{Supervisor. Supervision, Writing - Review \& Editing, and Administration.}

\author[a]{Ghazaleh Khodabandelou}[role=,
                                     suffix=]

\ead{ghazaleh.khodabandelou@u-pec.fr}
\ead[url]{https://orcid.org/ 0000-0002-8078-8461}
\credit{Methodology, Resources, and Visualization, Validation, and Writing - Review \& Editing.}

\author[b]{Zuheng Ming}

\ead{zuheng.ming@univ-paris13.fr}
\ead[url]{https://orcid.org/0000-0002-1094-3112}
\credit{Methodology, Resources, and Visualization.}

% Affiliations
\affiliation[a]{organization={Laboratoire Images, Signaux et Systémes Intelligents (LiSSi)–EA 3956, Université Paris-Est Créteil (UPEC)},
    addressline={122 Rue Paul Armangot, Vitry Sur Seine}, 
    city={Créteil},
    postcode={94010}, 
    country={France}}

\affiliation[b]{organization={Laboratoire L2TI, Institut Galilée, Université Sorbonne Paris Nord (USPN)},
    addressline={99 Avenue Jean-Baptiste Clément}, 
    city={Villetaneuse},
    postcode={93430}, 
    country={France}}

\cortext[cor1]{Corresponding author: Alice Othmani}

%\fntext[fn1]{This is the first author footnote. It is also common to the third author.}
%\fntext[fn2]{Another author footnote, this is a very long footnote, and it should be a long footnote. However, this footnote is not yet sufficiently long enough to make two lines of footnote text.}

%\nonumnote{This note has no numbers. In this work, we demonstrate the formation of a new type of polariton on the interface between a cuprous oxide slab and a polystyrene micro-sphere placed on the slab.}

% Abstract
\begin{abstract}
The integration of Artificial Intelligence (AI) and the Internet of Medical Things (IoMT) in healthcare, particularly through Machine Learning (ML) and Deep Learning (DL) techniques, has significantly advanced the prediction and diagnosis of chronic and terminal diseases. AI-driven models such as XGBoost, Random Forest, Convolutional Neural Networks (CNNs), and Long Short-Term Memory Recurrent Neural Networks (LSTM RNNs) have demonstrated remarkable accuracy, achieving over 98\% in predicting conditions like heart disease, chronic kidney disease (CKD), Alzheimer's, and lung cancer, relying on datasets sourced from public platforms like Kaggle and UCI, as well as private medical institutions and real-time IoMT sources.
Despite these achievements, significant challenges persist. The diversity in data quality, patient demographics, and data formats from various sources like hospitals and research initiatives poses integration challenges. Incorporating IoMT data, which is often vast and heterogeneous, adds further complexities, particularly in ensuring data interoperability and robust security measures to protect patient privacy. AI models often encounter overfitting, demonstrating high accuracy in controlled environments but struggling with new, unseen data in real-world clinical settings. Moreover, significant gaps exist in addressing multi-morbidity scenarios, particularly concerning rare and critical diseases like dementia, stroke, and various cancers. Future research must focus on advanced data preprocessing techniques to standardize and harmonize diverse data sources, enhancing data quality and interoperability. Techniques such as transfer learning, and ensemble methods are essential to improve model generalizability across different clinical settings. Addressing multi-morbidity in AI models requires a deeper exploration of disease interactions and the development of predictive models that consider chronic illness intersections. Additionally, creating standardized frameworks and open-source tools for integrating federated learning, blockchain, and differential privacy into IoMT systems is critical to ensure robust data privacy and security.

%\noindent Each keyword shall be separated by a \verb+\sep+ command.
\end{abstract}

\begin{graphicalabstract}
    \centering
    \includegraphics[width=\textwidth]{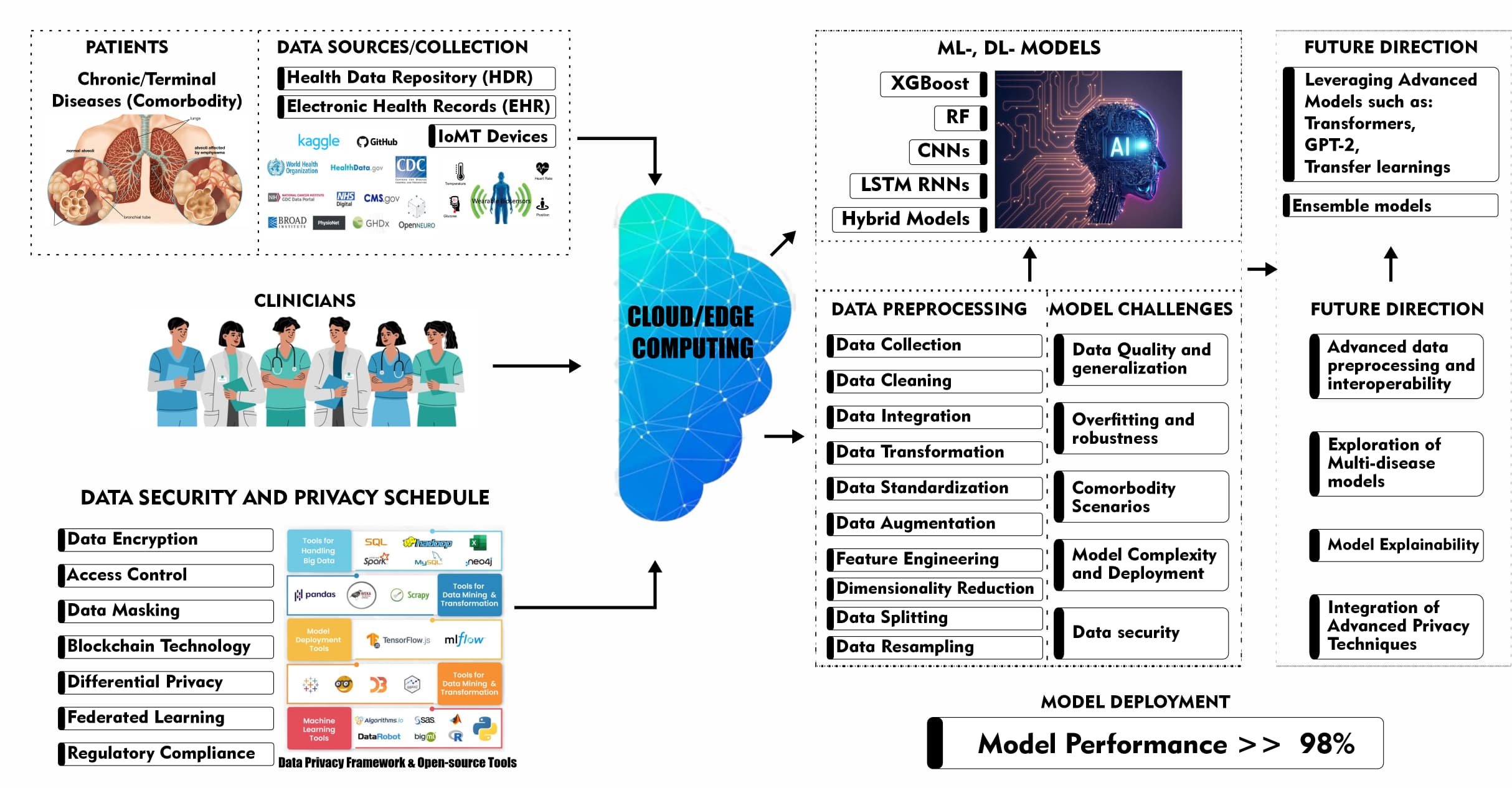}
\end{graphicalabstract}

\begin{highlights}
\item An overview of the background of prediction and detection of chronic and terminal diseases.
\item An assessment of data availability for AI-based IoMT chronic and terminal disease prediction.
\item An analysis of AI-IoMT-based models for chronic and terminal disease prediction.
\item An extensive study and analysis of ML-, DL-, and IoMT-based models used for chronic and terminal disease prediction.
\item A global discussion of methods, limitations, and valuable recommendations for future research.
\end{highlights}

% Keywords
% Each keyword is separated by \sep

\maketitle
\section{Introduction}
In the evolving landscape of healthcare technology, artificial intelligence (AI) and the Internet of Medical Things (IoMT) have emerged as transformative forces, revolutionizing the management of terminal diseases. These advancements enhance diagnostic accuracy, personalize treatment plans, improve patient monitoring, and optimize palliative care, significantly impacting patient outcomes and quality of life. Terminal diseases, such as cancer, advanced heart failure, and dementia, are incurable conditions that are likely to result in death \cite{marieWhatTerminalIllness2024,doppenCannabisPalliativeCare2022}. These diseases pose significant challenges due to their progressive nature and limited treatment options.
A disease that progresses to a terminal stage typically starts with mild symptoms that can initially be overlooked or attributed to other conditions. As the disease advances, these symptoms worsen, gradually impairing the patient's daily activities and overall health. Initially, treatment focuses on slowing disease progression and effectively managing symptoms. However, as the disease enters its later stages, including the terminal phase, curative treatments become less effective. Consequently, the focus shifts to palliative care, which aims to enhance the quality of life by alleviating symptoms such as pain, discomfort, and emotional distress \cite{amblas-novellasIdentifyingPatientsAdvanced2016,ekbergCommunicatingPatientsFamilies2021}. Advanced-stage chronic diseases, having progressed to severe stages with limited treatment options, often profoundly impact both the patient's quality of life and life expectancy, ultimately classifying them as terminal diseases \cite{reljicTreatmentTargetedUnderlying2017,farbickaSelectedAspectsPalliative2012}.
Advance-stage organ-based illnesses, such as End-Stage Renal Disease (ESRD) \cite{germinoWhenChronicIllness1998, powathilExperienceLivingChronic2023}, Chronic Obstructive Pulmonary Disease (COPD) \cite{valentiIdentificationPalliativeCare2023,tianHeterogeneousAgingMultiple2022, licherLifetimeRiskMultimorbidity2019}, and End-Stage Liver Disease (ESLD) \cite{antariantoInventingEngineeredOrganoids2022}, are also referred to as terminal diseases. They pose significant challenges due to their progressive nature and limited treatment options. Patients facing terminal illnesses often suffer from a heavy burden of symptoms, resulting in both physical and emotional distress \cite{jiaoConnectionsPhysicalPsychosocial2023, guayS25EmbracingSpiritual2023}.

Having established that chronic diseases at advanced stages can progress to become terminal illnesses, it is important to note that chronic diseases are the leading cause of death in the United States. Heart disease alone accounts for one-third of all fatalities, with cancer also contributing significantly to annual mortality rates \cite{siegelCancerStatistics20222022, hackerBurdenChronicDisease2024, moiniChapter12Chronic2023}. Cardiovascular diseases lead to the highest number of non-communicable disease (NCD) deaths, with 17.9 million fatalities each year \cite{whoNonCommunicableDiseases2023, mendisAddressingGlobalBurden2022}. They are followed by cancers, which cause 9.3 million deaths, chronic respiratory diseases, responsible for 4.1 million deaths, and diabetes, including kidney disease deaths due to diabetes, resulting in 2.0 million deaths. Collectively, these four categories of diseases contribute to over 80\% of all premature NCD deaths \cite{whoNonCommunicableDiseases2023}. Early prediction and detection are key strategies in managing these conditions effectively \cite{khalifaArtificialIntelligenceClinical2024, johnsonCostEffectivenessCase2021}.

By identifying symptoms early, healthcare providers can initiate timely interventions, personalized treatments, and supportive care measures to improve patient outcomes and enhance the quality of life of the patient \cite{amjadReviewInnovationHealthcare2023}. Enhancing end-of-life care for individuals with terminal illnesses necessitates a focus on employing a multidisciplinary approach involving algorithms and software applications to heighten early prediction and detection techniques, elevate patient outcomes, and mitigate healthcare expenses \cite{brigitteOpportunitiesComputationalTools2022, avatiImprovingPalliativeCare2018}. Moreover, IoMT datasets play an increasingly significant role in disease prediction, providing continuous streams of data on parameters such as heart rate, physical activity, glucose levels, and medication adherence \cite{umerHeartFailurePatients2023}.

The integration of Artificial Intelligence (AI) with the Internet of Things (IoT) and the Internet of Medical Things (IoMT) is revolutionizing healthcare delivery. IoMT connects medical devices, enabling seamless data collection, aggregation, and analysis. When combined with AI technologies, this interconnected network facilitates advanced decision-making within healthcare systems, leading to significant improvements in patient outcomes \cite{nigarIoMTMeetsMachine2023}. AI algorithms can analyze the vast amounts of data collected by IoMT devices to provide personalized and accurate diagnoses. For instance, wearable devices can continuously monitor vital signs, and AI can detect anomalies in real time, alerting healthcare providers to potential health issues before they become critical \cite{islamDeepLearningBasedIoT2023}. However, challenges accompany this innovative approach, including ensuring the availability of comprehensive datasets and addressing concerns regarding data privacy and security within IoMT systems \cite{netoSurveySecuringFederated2023}. Efforts to overcome these challenges are crucial to realize the full potential of AI and IoMT in revolutionizing healthcare delivery and ensuring safer, more efficient, and personalized patient care.

The scope of this systematic literature review focuses on advancements in the prediction of chronic and terminal diseases, particularly through the application of Machine Learning (ML) and Deep Learning (DL) techniques. We will explore the effectiveness of these methodologies in predicting diseases such as liver cancer (hepatocellular carcinoma - HCC), lung cancer, pancreatic cancer, dementia, advanced renal disease (ESRD), cardiovascular disease (CVD), and chronic obstructive pulmonary disease (COPD). Additionally, our review will explore how Internet of Medical Things (IoMT) technologies are used for data collection, aggregation, and analysis, while also addressing data privacy and security in IoMT implementation. Our ultimate goal is to offer insights into state-of-the-art approaches for predictive healthcare systems and to explore future innovations and directions within the field.

The contribution of this review paper addresses key issues in the field of chronic and terminal disease prediction using ML and DL models. It examines which chronic and terminal diseases have been effectively predicted, guiding future research and clinical applications. It also explores the datasets and sources used in predictive models, emphasizing the importance of data quality and availability for developing robust and accurate models. Additionally, the paper identifies the most promising ML and DL techniques for accurate and reliable predictions, informing the development of future predictive systems. Furthermore, it investigates the types of IoMT technologies employed and the privacy concerns associated with data collection and deployment, providing important insights into the infrastructure needed for efficient and secure implementation.

The structure of the remainder of this paper is organized as follows: 
Section.~\ref{Sect2} addresses the methodology of the survey. Section.~\ref{Sect3} discusses AI-IoMT- Based models specifically designed for predicting these conditions. Section.~\ref{Sect4} provides a global discussion of the survey results and findings. Section.~\ref{Sect5} identifies gaps in the current literature. Section.~\ref{Sect6} outlines future research directions. Finally, Section.~\ref{Sect7} presents the conclusions drawn from this study.

\section{Methodology} \label{Sect2}

\subsection{Background of the Study}
Early methods of disease prediction and detection relied heavily on clinical observation and basic diagnostic tools, such as physical examination and rudimentary laboratory tests. Healthcare providers assess patients' symptoms, medical histories, and visible signs of illness to make diagnoses, often based on their clinical experience and intuition \cite{vanstoneExperiencedPhysicianDescriptions2019, brushHowExpertClinicians2017}. In a scenario of hypertension, the health practitioner considers the individual's age, family history of hypertension, multiple monitored blood pressure readings, lifestyle, and high-sodium diet, and the person is identified as being at risk for hypertension \cite{onughaDietarySodiumPotassium2024}. However, these traditional approaches had significant limitations such as subjectivity, variability in diagnostic accuracy, and a lack of standardized protocols \cite{coskunBiasLaboratoryMedicine2024}. Challenges such as misdiagnosis, delayed diagnoses, and limited treatment options prompted advancements in diagnostic techniques and technologies \cite{dominguez-fernandezReviewTechnologicalChallenges2023}. 
Over time, laboratory assays have become more sensitive and efficient, allowing for rapid analysis of samples and the detection of disease-related biomarkers \cite{murphyRecentAdvancementsBiosensing2021,leeRapidDeepLearningassisted2024}. Similarly, advancements in medical imaging have led to detailed visualizations of internal structures, enhancing diagnostic accuracy and facilitating prompt intervention \cite{kleinVisualComputingMedical2009, kaseEffectiveThreedimensionalRepresentation2022}. Screening initiatives have broadened their focus to include populations at higher risk, employing advanced algorithms to identify diseases sooner and improve outcomes across various conditions \cite{kellyInterventionsImproveUptake2018}. Challenges affecting laboratory and scanning diagnostic methods, such as false positives, false negatives, and the demand for personalized health services, have prompted the adoption of AI technologies in clinical practice \cite{patelRevolutionizingDiseaseDiagnosis2024, hossamArtificialIntelligenceMedical2021}. These technologies provide improved accuracy, efficiency, and personalized care in disease detection and patient management \cite{pinto-coelhoHowArtificialIntelligence2023}.

However, despite advancements in healthcare technology, research into predicting and diagnosing chronic and terminal diseases confronts notable challenges, particularly regarding data quality and dimensionality. \cite{shelkeIoMTHealthcareDelivery2020, emmanuelSurveyMissingData2021}, device reliability and interoperability, and advanced algorithms \cite{padinjappurathugopalanEfficientPrivacyPreservingScheme2022}, the difficulty in symptom-based prediction \cite{ vijavashettySymptomBasedHealth2019} and the under-utilization of healthcare data due to privacy issues \cite{kheraInformationSecurityPrivacy2019}, challenges such as inference attacks, differential privacy, adversarial attacks, data poisoning, model and data tampering, and compliance with data protection regulations make safeguarding data privacy and security in AI applications essential, thereby reducing the availability of datasets \cite{aldoseriReThinkingDataStrategy2023}. Similarly utilizing IoMT devices to detect and diagnose chronic diseases presents significant data privacy and security challenges. These devices collect sensitive patient data, making them vulnerable to cyber-attacks that could compromise patient confidentiality \cite{ ksibiIoMTSecurityModel2023}. Implementing robust data security measures is essential to protect patient information and ensure the reliability and safety of chronic and terminal disease diagnoses via IoMT devices \cite{gallosMedSecuranceProjectAdvanced2023, samuelAnonymousIoTBasedEHealth2023}.

\subsection{Data Availability for AI-Based IoMT Chronic and Terminal Disease Prediction} 

Datasets essential for AI-based disease prediction are available in both public and private repositories, serving as fundamental resources for IoMT system research and development. However, datasets specifically tailored for IoMT applications are typically inaccessible to the public due to stringent privacy regulations that govern medical data. For instance, regulations such as the General Data Protection Regulation (GDPR) in Europe and the Health Insurance Portability and Accountability Act (HIPAA) in the United States impose strict guidelines on the collection, storage, and sharing of sensitive health information \cite{schmidtRegulatoryChallengesHealthcare2020,bakareDATAPRIVACYLAWS2024}. These regulations prioritize patient confidentiality and aim to prevent unauthorized access or misuse of personal health data. As a result, researchers and developers often work with anonymized or aggregated datasets to comply with these regulations while still advancing IoMT technologies and applications \cite{hirecheSecurityPrivacyManagement2022, andersonSyntheticDataGeneration2014}.

Meanwhile, publicly accessible datasets relevant to IoMT or medical IoT are available on platforms such as Kaggle, UCI, and through public institutions, providing valuable resources for advancing IoMT technologies. For instance, Rash et al. (2022) utilized data from UCI sources to develop a predictive model for diseases such as breast cancer, diabetes, heart attack, hepatitis, and kidney disease. They employed Particle Swarm Optimization (PSO) to extract features from these datasets and achieved high accuracy rates with their ANN model, with 98.23\% for breast cancer and 98.90\% for kidney disease, surpassing other algorithms tested \cite{rashidAugmentedArtificialIntelligence2022} while private datasets typically originate from medical centers, hospitals, research institutions, and occasional collaborations with private entities. 
%\cite{woodsMedicalDigitalRepositories2002}
David et al. (2021) utilized federated learning for distributed model training, leveraging a private dataset of 600,000 patient records to develop a deep learning-based multilayer convolutional neural network (CNN) for early chronic illness detection, achieving high accuracy, low error rates, and high precision \cite{daidEffectiveMechanismEarly2021}.

Integrating data from multiple and diverse sources presents significant challenges such as data inconsistency, heterogeneity, unstructured formats, quality issues, semantic discrepancies, scalability concerns, privacy and security risks, real-time integration difficulties, and high costs and resource requirements \cite{batistaDataMultipleWeb2020}. To data consistency a uniform data integration approach is essential, followed by data preprocessing to address issues such as missing or duplicate data, normalization, and data accuracy. Certain diseases require specialized datasets for accurate prediction, driven by factors such as disease complexity, rarity, or the need for specific data modalities. For instance, chronic conditions often necessitate longitudinal datasets that track patients' health outcomes over extended periods. The Cancer Imaging Archive (TCIA) exemplifies this need as it houses a vast repository of cancer-related imaging and associated data for the U.S. National Cancer Institute. With 30.9 million radiology images from about 37,568 subjects, organized by tumor type, TCIA provides comprehensive resources including analytic results and clinical data \cite{priorPublicCancerRadiology2017}. Another is the National Institutes of Health, NIH dataset containing 82 abdominal CT scans focused on the pancreas, enhanced with contrast for improved visualization and featuring a resolution of 512 × 512 pixels. It includes scans from 53 male and 27 female subjects, aged between 18 and 76 years. Among them are 17 healthy kidney donors and additional patients without pancreatic lesions. For precise labeling, a medical student manually annotated each CT slice under the guidance of an experienced radiologist \cite{rothNIHPancreasCTDataset2017}. The Alzheimer's Disease Neuroimaging Initiative (ADNI) database is designed to identify and track the progression of Alzheimer's disease (AD) employing various data collection modalities, including MRI for structural brain imaging, PET for metabolic and amyloid plaque imaging, clinical and neuropsychological assessments for cognitive evaluation, biomarker analyses from cerebrospinal fluid and blood samples, and genetic data. This comprehensive dataset facilitates early diagnosis, biomarker discovery, and pathophysiological studies while enhancing clinical trial methodologies \cite{muellerWaysEarlyDiagnosis2005}. Similarly, OASIS-3 is a longitudinal, multimodal dataset on normal aging and Alzheimer’s disease from WUSTL Knight ADRC, spanning 30 years. It includes 1,378 participants aged 42-95, with 755 cognitively normal and 622 cognitively impaired individuals. The dataset offers 2,842 MRI sessions, over 2,157 PET scans with multiple tracers, and 451 Tau PET sessions, all anonymized with normalized dates \cite{lamontagneOASIS3LongitudinalNeuroimaging2019}.

In addition, the MIMIC-III database is a comprehensive, single-center repository containing detailed patient information from critical care units at a large tertiary care hospital. The database is openly accessible and includes a wide range of data such as vital signs, medications, laboratory results, care provider observations and notes, fluid balance, procedure and diagnostic codes, imaging reports, hospital length of stay, and survival data. \cite{johnsonMIMICIIIFreelyAccessible2016}. Furthermore, \cite{bilionisSurveyPublicData2023} unveiled approximately forty-eight public databases on biomedical datasets associated with chronic illnesses, describing each dataset's characteristics such as therapeutic areas covered, sample size, publication year, and available features like socio-demographic, clinical, physical activity, time-series, and psychometric data. Additionally, the report highlights whether the dataset is stored in an archival repository with a Digital Object Identifier (DOI), if there is a related publication of its utilization, and if comprehensive ethics statements and anonymization details are included in the data release. Further points of interest include the accessibility of the datasets, potential usage restrictions, data quality assessments, and any collaboration opportunities offered through the datasets. Extractions from these datasets and other relevant dataset sources provide comprehensive insights into publicly available biomedical information for chronic diseases including cancer, diabetes, heart diseases, brain disorders, and more across a wide spectrum. This initiative aims to raise awareness about the wealth of resources available for studying chronic diseases, as outlined in Table \ref{tab:used_datasets}. providing details such as year of release, disease area, dataset name, participant numbers, data type, and sources for each study.

\subsubsection{Comparative Analysis of Predictive Performance of Selected Datasets}
\paragraph{Alzheimer’s disease prediction using ADNI-MRI datasets}
The progression and diagnosis of Alzheimer's Disease (AD) have been enhanced through imaging modalities and machine learning. A study using MRI data showed that analyzing the hippocampus (H), entorhinal cortex (EC), and middle temporal cortex (MTC) provided strong predictive performance, with the EC as the most effective predictor, achieving AUC values of 0.86, 0.85, and 0.82 for 1, 2, and 3-year predictions, respectively \cite{liEarlyPredictionAlzheimer2020}. Another study combined MRI and PET data with discrete wavelet transform (DWT) and VGG16, using a vision transformer to classify fused images, achieving 81.25\% and 93.75\% accuracies for MRI and PET, respectively \cite{odusamiPixelLevelFusionApproach2023}. A comparative analysis proposed an ensemble model (XGB + DT + SVM) that achieved 95.75\% accuracy \cite{khanEnsembleModelDiagnostic2022}. Additionally, a CGAN model using MRI and PET data improved classification accuracy to 94\% by extracting and fusing structural and metabolic features\cite{choudhuryCoupledGANArchitectureFuse2024}.
Table \ref{tab:alzheimer_ADNI}. shows the summary of these approaches to Alzheimer’s disease prediction using ADNI-MRI datasets.

\begin{table}[ht]
\caption{Summary of  approaches to Alzheimer’s disease prediction using ADNI-MRI datasets. Legend: Sensitivity (Sens.) ;  Specificity (Spec.) ; Acurracy (Acc.) ;
Area under curve (AUC.)}
\label{tab:alzheimer_ADNI}
\centering
\begin{tabular}{p{2cm}p{2.5cm}p{2.5cm}}

\hline
\textbf{Ref.}  & \textbf{Approach} & \textbf{Evaluation} \\
\hline
Li et al. (2020) \cite{liEarlyPredictionAlzheimer2020}  &  Principal Analysis through Conditional Expectation (PACE)  & Sens.: 80\%, Spec.: 70\%, Acc.: 75\%, AUC: 80\% \\
\hline
Khan et al. (2022) \cite{khanEnsembleModelDiagnostic2022}  & (XGB+DT+SVM) & Acc.: 95.75\% \\
\hline
Odusanmi et al. (2023) \cite{odusamiPixelLevelFusionApproach2023}
 & VGG16 + inverse discrete wavelet transform (IDWT) & Acc. 81.25\% \\
\hline

Choudhury et al. (2024) \cite{choudhuryCoupledGANArchitectureFuse2024}  & Coupled GAN & Acc.: 94.00\% \\
\hline
\end{tabular}
\end{table}

\paragraph{Alzheimer’s disease prediction using OASIS 3-MRI datasets.}
Recent studies on Alzheimer's disease (AD) detection using OASIS 3-MRI data have explored various advanced methods to improve accuracy and reliability. One approach combines CNN, RNN, and LSTM models with Bagging to enhance performance, achieving an accuracy of 92.22\% compared to 89.75\% without Bagging \cite{duaCNNRNNLSTM2020}. Another study employs an ensemble voting method, surpassing traditional techniques with an impressive 96.4\% accuracy and 97.2\% AUC on the OASIS dataset \cite{chatterjeeVotingEnsembleApproach2022}. A third study focuses on CNNs integrated with Transfer Learning and Generative Adversarial Networks (GANs), demonstrating significant accuracy improvements of up to 40.1\% over existing methods \cite{chuiMRIScansBasedAlzheimer2022}. Additionally, a novel approach combines clinical, MRI segmentation, and psychological data using nine machine learning models, with Random Forest achieving the highest performance at 98.81\% accuracy, and employs SHAP for model explainability \cite{jahanExplainableAIbasedAlzheimer2023}.
Table \ref{tab:alzheimer_OASIS}. shows summary of these approaches to Alzheimer’s disease prediction using OASIS 3-MRI datasets.

\begin{table}[ht]
\centering
\caption{Summary of approaches to Alzheimer’s disease prediction using OASIS-3 MRI datasets. Legend: Sensitivity (Sens.) ;  Specificity (Spec.) ; Acurracy (Acc.) ;
Area under curve (AUC.)}
\label{tab:alzheimer_OASIS}
\begin{tabular}{p{2cm}p{2.5cm}p{2.5cm}}
\hline
\textbf{Ref.}  & \textbf{Method} & \textbf{Evaluation} \\
\hline
Dua et al. (2020) \cite{duaCNNRNNLSTM2020} & SVM, CNN, RNN, LSTM & Acc.: 89.75\%, 92.22\% \\
\hline
Chatterjee and Byun (2022) \cite{chatterjeeVotingEnsembleApproach2022} & VGG16 & Acc.: 96.4\%, AUC: 97.2\% \\
\hline
Chui et al. (2022) \cite{chuiMRIScansBasedAlzheimer2022} & GAN & Acc.: 94\%, 93\%, 95\% \\
\hline
S. Jahan et al. (2023) \cite{jahanExplainableAIbasedAlzheimer2023}  & RF, LR, DT, MLP, KNN, GB, AdaB, SVM, and NB & Acc.: 94.51\% to 98.94\% \\
\hline
\end{tabular}
\end{table}

\paragraph{Liver disease prediction using UCI Indian Liver Patient dataset (ILPD).}

Advancements in liver disease prediction have leveraged machine learning techniques to enhance diagnostic accuracy. One study applied various classification algorithms using WEKA on liver patient data, achieving a maximum accuracy of 74.2\% \cite{muthuselvanClassificationLiverPatient2018}. Another model focused on predicting the probability of liver disease progression using logistic regression, achieving 72.4\% accuracy, 90.3\% sensitivity, and 78.3\% specificity with the ILPD dataset \cite{abdalradaPredictiveModelLiver2019}. A third study compared SVM and logistic regression, noting that SVM achieved 88\% accuracy, though both algorithms showed decreased performance with PCA and SMOTE preprocessing \cite{barusLiverDiseasePrediction2022}. Additionally, a novel approach utilized ensemble learning and advanced preprocessing methods on the ILPD dataset, with the Extra Tree classifier and Random Forest achieving high accuracies of 91.82\% and 86.06\%, respectively \cite{mdEnhancedPreprocessingApproach2023}. Finally, a comprehensive evaluation of twelve machine learning algorithms on the BUPA Liver Disease dataset found Decision Tree to be the most effective, with an accuracy of 86.67\% \cite{kumarLiverDiseasePrediction2023}. Table \ref{tab:liver_disease_prediction}. shows summary of approaches to liver disease prediction using UCI Indian Liver Patient dataset (ILPD).

\begin{table}[ht]
\centering
\small
\caption{Summary of approaches to liver disease prediction using UCI Indian Liver Patient dataset (Multivariate-Indian Liver Patient Data (LIPD)). Legend: Sensitivity (Sens.) ;  Specificity (Spec.) ; Precision (Prec.) ; Acurracy (Acc.) ;
Area under curve (AUC.)}
\label{tab:liver_disease_prediction}
\begin{tabular}{p{2cm}p{2.5cm}p{2.5cm}}
\hline
\textbf{Ref.}  & \textbf{Method} & \textbf{Evaluation} \\
\hline
M. et al. (2018) \cite{muthuselvanClassificationLiverPatient2018} & NB, K-star, J48-decision tree, Random Tree & RT:74.2\% \\
\hline
A. et al. (2019) \cite{abdalradaPredictiveModelLiver2019}  & LR & Acc.:72.4\%, Sens.:90.3\%, Spec.:78.3\%, ROC:0.758\% \\
\hline
Barus et al. (2022) \cite{barusLiverDiseasePrediction2022}  & SVM,LR (PCA and SMOTE) & Before PCA and SMOTE:  SVM: acc.88\%  After SVM: acc.87\%  \\
\hline
 Md et al. (2023) \cite{mdEnhancedPreprocessingApproach2023}   &   GB, XGB, Bagging, RF, Extra Tree, and Stacking & Extra Tree:91.82\%, RF:86.06\% test acc. \\
\hline
Kumar et al. (2023) \cite{kumarLiverDiseasePrediction2023}  & ML Classifiers & DT acc.86.67\%, prec., recall, F1:0.87, 0.87, and 0.86 \\
\hline
\end{tabular}
\end{table}

\paragraph{Pancreas detection using NIH Pancreas-CT dataset.}

Recent advancements in pancreas segmentation and pancreatic cancer prediction employ various innovative techniques. One CAD model uses Simple Linear Iterative Clustering (SLIC) for pancreas segmentation on CT images, Dual Threshold Principal Component Analysis (DT-PCA) for feature extraction, and a hybrid Feedback-Support Vector Machine-Random Forest (HFB-SVM-RF) model for classification, achieving 96.47\% accuracy and high sensitivity and specificity \cite{liEffectiveComputerAided2018}. Another study developed an artificial neural network (ANN) using data from over 800,000 respondents, achieving a sensitivity of 87.3\% and specificity of 80.8\% \cite{muhammadPancreaticCancerPrediction2019}. A method for automatic pancreas segmentation in MRI and CT scans achieved a Dice Similarity Coefficient (DSC) of 79.3\% to 81.6\% through a three-stage process including contrast enhancement and 3D segmentation \cite{asaturyanMorphologicalMultilevelGeometrical2019}. Lastly, the DSD-ASPP-Net model, which integrates coarse segmentation with image context, achieved an average DSC of 85.49\% \cite{huAutomaticPancreasSegmentation2021}. Table \ref{tab:pancreas-CT}. shows summary of approaches for pancreas segmentation/detection using NIH Pancreas-CT dataset.

\begin{table}[ht]
\centering
\small
\caption{Summary of approaches for pancreas segmentation/detection using NIH Pancreas-CT dataset.}
\label{tab:pancreas-CT}
\begin{tabular}{p{2cm}p{2.5cm}p{2.5cm}}
\hline
\textbf{Reference} & \textbf{Method} & \textbf{Evaluation} \\
\hline
Siqi Li et al. (2018) \cite{liEffectiveComputerAided2018}  & Dual threshold-PCA & Dice coefficient.78.9\%; Jaccard index 65.4\% \\
\hline
M. et al. (2019) \cite{muhammadPancreaticCancerPrediction2019}  & ANN & Acc.:87.3\% and 80.7\%, Spec.:80.8\% and 80.7\% \\\hline
Hykoush Asaturyan et al. (2019) \cite{asaturyanMorphologicalMultilevelGeometrical2019} & Cont. Max-Flow,Min-Cuts,Structured FED & DSC 79.3 ± 4.4\% \\
\hline
Peijun Hu et al. (2021) \cite{huAutomaticPancreasSegmentation2021} & Distance-Based Saliency with Dense Atrous Spatial Pyramid Pooling (DenseASPP) &  Dice-Sørensen Coefficient (DSC) 85.49±4.77\% \\
\hline

\hline
\end{tabular}
\end{table}

\subsection{Development of Search Strategy}

To conduct this survey paper, a thorough literature search was conducted using Google Scholar, Semantic Scholar, and ResearchGate databases. The search employed specific keywords and phrases using related terms and synonyms with Boolean operators (AND, OR, NOT) to combine keywords such as "AI", "IoMT", "Secure", "Deep-Learning", "Machine Learning", "Advanced Chronic Diseases", "Terminal Disease", "Prediction", and "Detection" effectively. Some of the phrases used in the search on the databases are "AI-based model for Terminal and Chronic Disease Prediction", "IoMT and AI-based model for Terminal and Chronic Disease Prediction", "Secure AI-based model for Terminal and Chronic Disease Prediction", "AI-based model for Multiple Chronic Disease Prediction", "IoMT AI-based model for Multiple Chronic Disease Prediction", "Secure AI-based model for Chronic Disease Prediction", and "AI-based model for Terminal Disease Prediction". The most effective keywords for the search are "AI", "IoMT", "Chronic Disease", and "Prediction". Additionally, filters were applied to focus on journal articles, conference papers, and reviews within relevant subject areas such as Computer Science, Medicine, and Engineering.

Furthermore, various recent literature surveys on AI and IoMT applications in chronic disease prediction and management are assessed highlighting several critical gaps in each of the related survey papers. Souza-Pereira et al.'s (2020) survey paper lacks focus on major chronic and terminal diseases in clinical decision support systems. Similarly, Xie et al.'s (2021) paper lacks broader inclusion of key ML techniques and other chronic diseases. Ahsan et al.'s (2022) paper does not sufficiently integrate the Internet of Medical Things (IoMT) with ML-based diagnostic systems, particularly for major chronic and terminal diseases. Anilkumar et al.'s (2023) work lacks the utilization of advanced AI techniques and IoMT in multi-chronic disease prediction. Ajagbe et al.'s (2024) paper lacks the application of IoMT and deep learning techniques to both pandemics and chronic diseases. Lastly, Merabet et al.'s (2024) paper lacks detail on the integration of data privacy and security in AI and IoMT applications. These gaps highlight the necessity of our survey paper to fill these gaps. Table \ref{tab:survey_papers}. shows summary of related survey papers showing references, focus, and identified gaps.

\begin{table}[h]
\centering
\caption{Summary of related survey papers showing references, focus, and identified gaps}
\label{tab:survey_papers}
\begin{tabular}{p{1.5cm}p{2.5cm}p{2.8cm}}
\hline
\textbf{Reference} & \textbf{Focus} & \textbf{Gap Identified} \\ \hline
Souza-Pereira et al. (2020) \cite{souza-pereiraClinicalDecisionSupport2020} & Disease management, personalized guidance & IoMT and major diseases not focused  \\ \hline
Xie et al. (2021) \cite{xieMultiDiseasePredictionBased2021} & Deep learning for diseases & Key ML techniques and IoMT-privacy not focused  \\ \hline

Shamout et al. (2021) \cite{shamoutMachineLearningClinical2021} & clinical outcome prediction models  & Chronic Diseases and IoMT-privacy not focused  \\ \hline

Qayyum et al. (2021) \cite{qayyumSecureRobustMachine2021} & privacy-preserving ML for healthcare applications  & Chronic Diseases and Model Performance not focused \\ \hline

Ahsan et al. (2022) \cite{ahsanMachineLearningBasedDiseaseDiagnosis2022} & ML-based diagnosis diseases & IoMT-privacy and major chronic diseases not focused  \\ \hline
Anilkumar et al. (2023) \cite{anilkumarMultiChronicDisease2023} & AI techniques for diseases & IoMT-privacy and other AI techniques not focused  \\ \hline
Ajagbe et al. (2024) \cite{ajagbeDeepLearningTechniques2024} & DL for pandemics, particularly Covid-19 & IoMT-privacy and major chronic diseases not focused \\ \hline
Merabet et al. (2024) \cite{merabetMultipleDiseasesForecast2024} & Forecasting disease through AI and IoMT & data privacy and security measures not focused  \\ \hline
\end{tabular}
\end{table}

\subsection{Eligibility and Inclusion Criteria}

This survey is performed by respecting the PRISMA-CI \cite{zhangIntroductionPRISMACIExtension2019}, an extension of PRISMA that reports guidelines about systematic reviews.We have followed guidelines to collect our papers. The eligibility and inclusion criteria were carefully designed to focus on open-access journals, chapters, conference articles, and publications that specifically addressed the interest area, utilizing targeted keywords and phrases such as AI or IoMT-based papers on terminal or chronic disease prediction. Comprehensive searches were conducted across prominent academic platforms including Google Scholar, Semantic Scholar, and Research Gate, yielding substantial results with 12,880, 12,500, and 5,189 items respectively. To ensure relevance and quality, further filters were applied to refine the search results, including limiting publication dates from 2015 to 2024 and prioritizing reputable scientific journals and conference proceedings hosted by publishers such as IEEE Xplore, MDPI, Elsevier, PubMed, Springer, Frontiers, and Hindawi. Additionally, the search extended beyond academic literature to encompass authoritative reports from sources like WHO, CDC, and mariecurie.org.uk. These sources provided essential statistics and definitions crucial for framing the context of chronic and terminal diseases within the scope of AI and IoMT applications.

\subsection{Screening}

A total of 6,327 articles were identified the databases. Of these, 2669 were duplicates; these duplicates were identified automatically using Mendeley and Zotero duplicate detection, along with some manual methods of duplicate removal such as visual inspection of titles, abstracts, and author lists, and looking for citations within the articles to identify potential duplicates were employed during the screening process. Additionally, 3,087 articles were removed for other reasons, such as irrelevance to the research topic, non-English language, and non-peer-reviewed sources. This left 566 articles for screening, which were relevant to the research question, publication type, and study design. During the screening, 13 articles could not be assessed due to inaccessible full texts and lack of response from authors within the study period. Furthermore, 486 articles did not meet our inclusion criteria based on the keywords "Prediction and Detection" and "Chronic or Terminal Disease". Ultimately, 67 full-text papers met the primary inclusion criteria, with 20 full-text papers on the final inclusion criteria focusing on AI-IoMT-based models for predicting single or multiple chronic and terminal diseases.

\begin{figure*}[htbp]
    \centering
    \includegraphics[width=\textwidth]{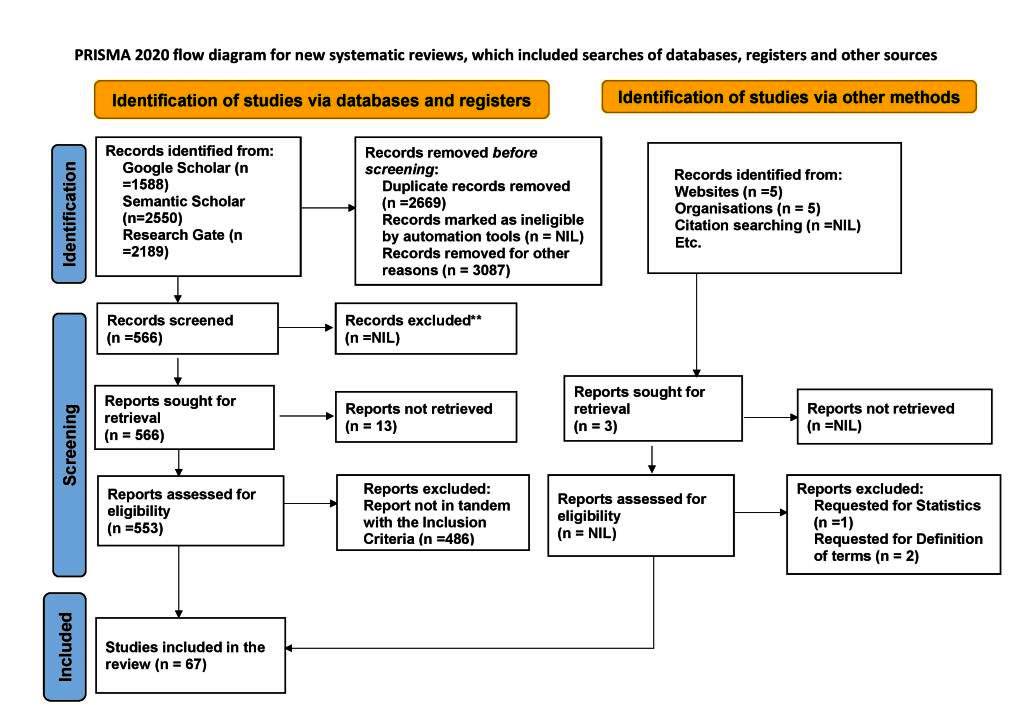}
    \caption{Flowchart adapted from \cite{pagePRISMA2020Statement2021} illustrating the PRISMA-based selection process for the 67 research articles that met our inclusion criteria, in alignment with the PRISMA guidelines.}
    \label{prisma_new.jpg}
\end{figure*}

\begin{figure}[htbp]
    \centering
    \begin{subfigure}[b]{.5\textwidth}
        \centering
        \includegraphics[width=\textwidth]{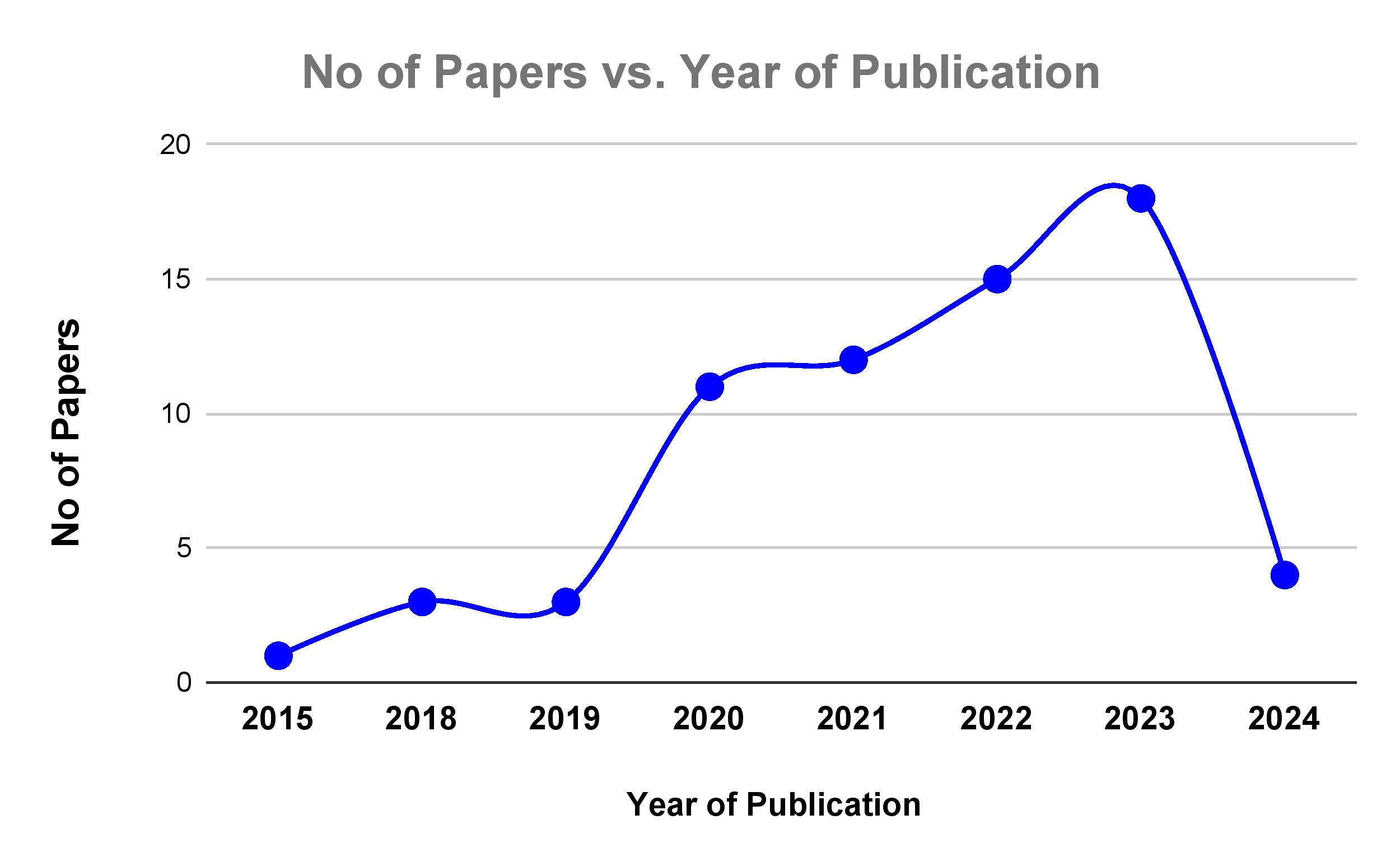}
        \caption{Line chart showing number per year of publication}
        \label{fig:years_new}
    \end{subfigure}
    \hfill
    \newline
    \begin{subfigure}[b]{.5\textwidth}
        \centering
        \includegraphics[width=\textwidth]{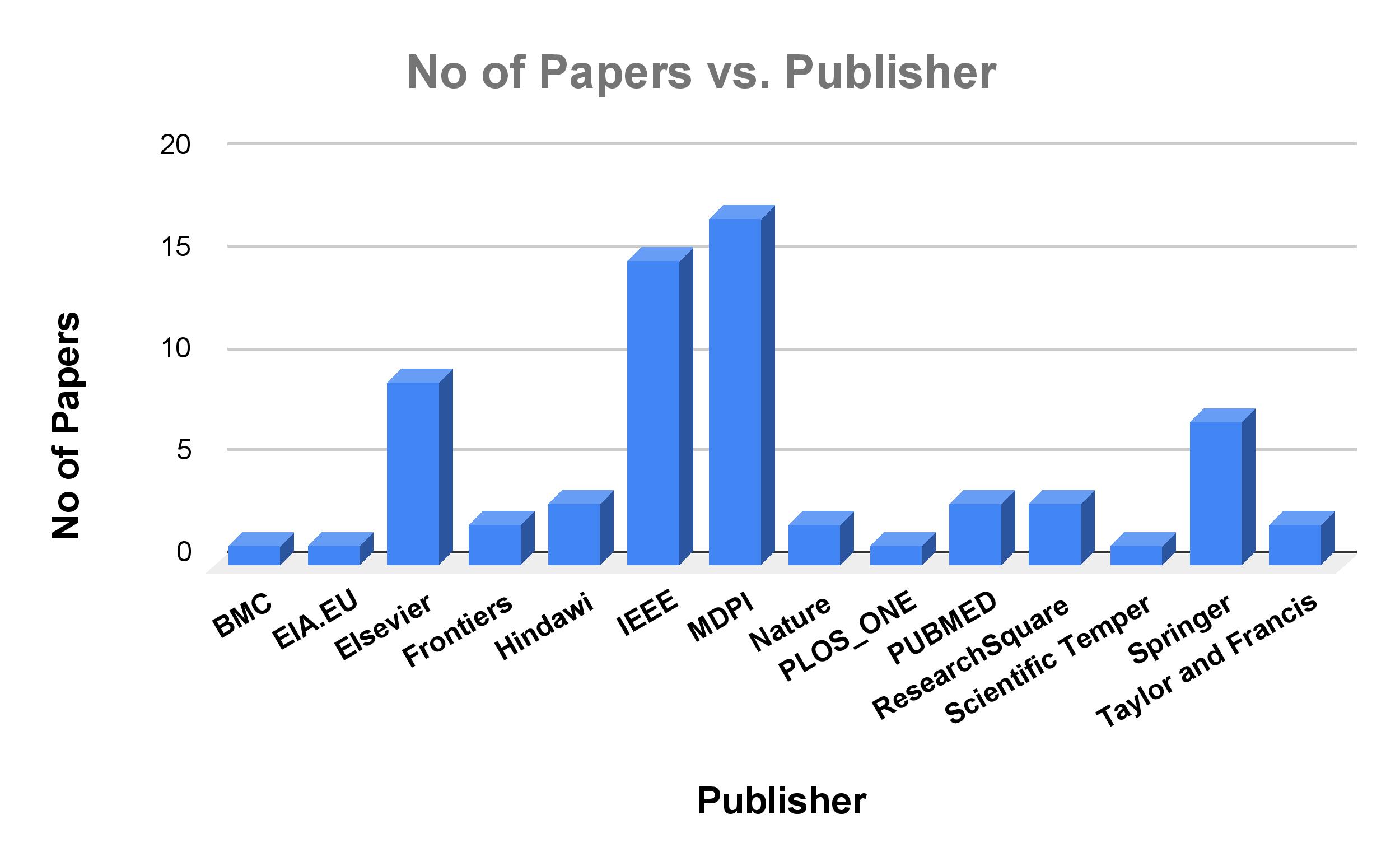}
        \caption{Bar chart showing number of reports per publisher}
        \label{fig:publishers_new}
    \end{subfigure}
    \caption{Showing year of publication and name of publisher of each report}
    \label{fig:publications}
\end{figure}

\begin{figure}[!ht]
    \centering
    \includegraphics[width=\columnwidth]{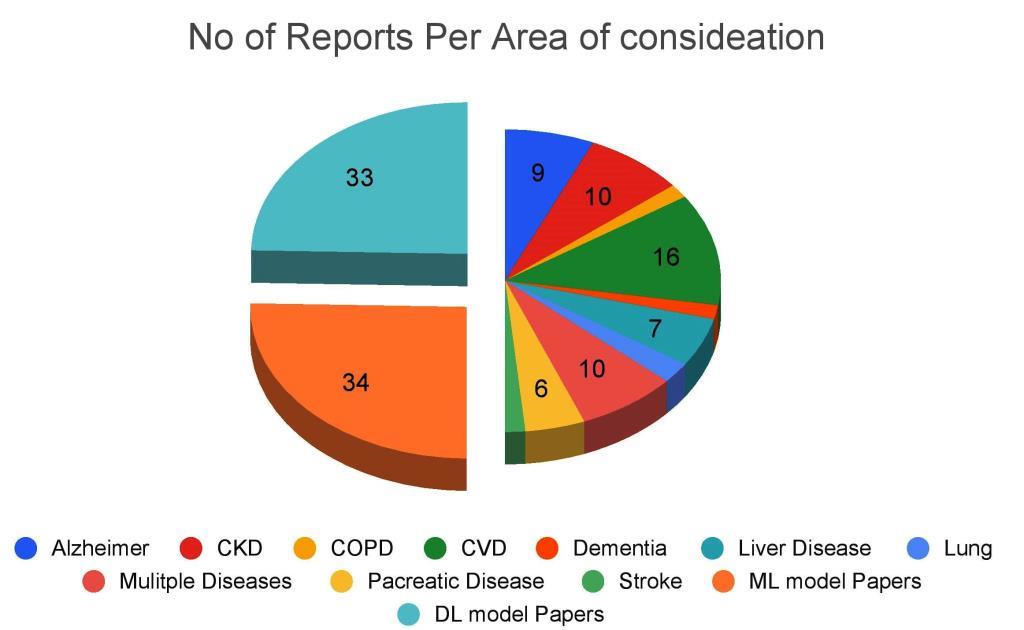}
    \caption{The number of papers reporting on each area of the disease under consideration, as well as the distribution of papers based on ML and DL models.}
    \label{fig:num_papers}
\end{figure}

The flow diagram, adapted from the PRISMA 2020 Statement \cite{pagePRISMA2020Statement2021}, is illustrated in Figure \ref{prisma_new.jpg}. Additionally, Figure \ref{fig:publications} presents the 67 papers categorized by year of publication and publisher. Furthermore, Figure \ref{fig:num_papers} summarizes the distribution of papers across different disease areas and ML and DL models. It shows that cardiovascular disease (CVD) is the most frequently covered disease with 16 papers, followed by chronic kidney disease (CKD) and multiple diseases, each with 10 papers. Alzheimer’s disease and liver disease have 9 and 7 papers, respectively. Other conditions, such as pancreatic disease, lung disease, stroke, and dementia, have fewer papers, ranging from 2 to 6. Additionally, 34 papers use machine learning (ML) models, and 33 papers use deep learning (DL) models.

\section{AI-IoMT-Based Model for Predicting Chronic and Terminal Diseases} \label{Sect3}

This section explores the significant influence of ML, DL, and IoMT on healthcare, focusing on disease prediction and management. It encompasses conditions such as dementia, Alzheimer's disease, cardiovascular disease (CVD), chronic kidney disease (CKD), end-stage renal disease (ESRD) , and liver, lung, and pancreatic diseases \cite{whoNonCommunicableDiseases2023, mendisAddressingGlobalBurden2022}. Integrating these algorithms within the framework of the Internet of Medical Things (IoMT) underscores the creation of accurate predictive models while emphasizing data privacy and security measures. In the Machine Learning approaches section, essential ML techniques such as XGBoost and Random Forest are central for precise disease prediction and management. Ensemble methods like Gradient Boosting and Bagging further enhance predictive accuracy by combining multiple ML models, thereby increasing reliability across diverse healthcare datasets. Conversely, the Deep Learning section highlights the transformative impact of DL models such as Convolutional Neural Networks (CNNs), Recurrent Neural Networks (RNNs), and variations like Long Short-Term Memory (LSTM). CNNs excel in extracting features from imaging technologies such as MRI and CT scans, while RNNs analyze time-series data to monitor disease progression and detect early indicators. The integration of IoMT explores their combined potential in enabling real-time monitoring and personalized healthcare solutions. Robust encryption protocols and secure data transmission mechanisms ensure patient privacy and confidentiality, addressing ethical concerns inherent in healthcare data management.

\subsection{ML Techniques Deployed for Chronic and Terminal Disease Prediction}

Disease prediction research has advanced significantly by integrating cutting-edge machine learning (ML) algorithms, sophisticated feature extraction methods, and robust privacy measures, ensuring high accuracy and secure data handling across diverse healthcare datasets. This comprehensive exploration covers cardiovascular diseases (CVD), chronic kidney disease (CKD) and end-stage renal disease (ESRD), dementia and Alzheimer's disease, liver, lung, and pancreatic diseases, as well as multiple diseases. CVD prediction employs advanced ML techniques like Bagging-Fuzzy-GBDT and Spark MLlib within Internet of Medical Things (IoMT) infrastructures, providing high accuracy while maintaining robust data privacy. CKD/ESRD prediction utilizes algorithms such as XGBoost and ANFIS for precise identification of disease stages and risk factors. Dementia and Alzheimer's prediction leverages ML models like Random Forest and Fuzzy Rule-Based Systems, using neuroimaging and wearable sensor data for high accuracy. Liver, lung, and pancreatic disease prediction uses algorithms like XGBoost and Federated Deep Extreme Machine Learning, demonstrating superior sensitivity, specificity, and generalizability. Multiple disease prediction integrates optimized techniques such as ANN-PSO and Multilayer Perceptron, achieving exceptional accuracy and efficiency.

In the following subsections, we will provide a detailed analysis of each family of chronic and terminal diseases prediction using machine learning models

\subsubsection{CVD (Heart Disease) Prediction}

The landscape of heart disease prediction research showcases advanced methodologies, diverse datasets, and critical privacy measures to improve accuracy and data security in healthcare. Each study offers unique insights and innovations, shaping the future of personalized medicine. Zhang et al. (2018) introduced a Privacy-Preserving Disease Prediction (PPDP) scheme using Single-Layer Perceptron (SLP) classifiers, securing medical data in cloud environments while achieving strong predictive performance \cite{zhangPPDPEfficientPrivacypreserving2018}. Adewole et al. (2021) developed an IoMT Cloud-based Personalized Healthcare Model (CPHM) integrating SVM, Naïve Bayes, and KNN on a large-scale Kaggle dataset, emphasizing scalability and data efficiency with robust security measures like digital watermarking \cite{adewoleChapterFiveCloudbased2021}.

Yuan et al. (2022) presented an AI-Based Heart Disease Prediction Model using the Bagging-Fuzzy-GBDT algorithm, demonstrating high accuracy and privacy strategies within IoMT infrastructures \cite{yuanStableAIBasedBinary2022}. Ed-daoudy et al. (2024) utilized Spark MLlib and algorithms like RF and SVM in a scalable architecture supporting real-time data processing and secure patient data handling \cite{ed-daoudyScalableRealtimeSystem2023}. Asif et al. (2023) explored ensemble learning techniques with Extra Tree Classifier, RF, XGBoost, and CatBoost, enhancing heart disease prediction through rigorous model optimization and validation \cite{asifEnhancingHeartDisease2023}. Ogunpola et al. (2024) achieved significant accuracy using ML classifiers expecailly with  XGBoost on datasets from Mendeley and Kaggle, emphasizing privacy measures like encryption and differential privacy in IoMT for real-time health monitoring \cite{ogunpolaMachineLearningBasedPredictive2024}.

\subsubsection{CKD/ESRD Prediction}
 
The prediction of kidney disease stages and associated risk factors has been extensively explored using various data mining and machine learning algorithms. E. A. Radya and A. S. Anwar (2019) employed multiple algorithms, including Probabilistic Neural Networks (PNN), Multilayer Perceptron (MLP), Support Vector Machine (SVM), and Radial Basis Function (RBF), to predict chronic kidney disease (CKD) stages. Their study, using a dataset of 361 CKD Indian patients from the UCI machine learning repository, found that the PNN algorithm exhibited particularly high accuracy across different CKD stages  \cite{radyPredictionKidneyDisease2019}. Md. A. Islam et al. (2020) identified risk factors for CKD using algorithms such as Decision Tree (DT), Naïve Bayes (NB), Random Forest (RF), and Linear Regression (LR). Their research highlighted that the Random Forest algorithm achieved the highest accuracy at 98.8858\%, with hemoglobin identified as a significant risk factor for CKD \cite{islamRiskFactorPrediction2020}. 

In a large-scale study, Z. Segal et al. (2020) used the XGBoost algorithm to predict end-stage renal disease (ESRD) risk from over 20 million medical insurance claims from a major U.S. health insurance company. The XGBoost model demonstrated high performance, achieving a C-statistic of 0.93, sensitivity of 0.715, and specificity of 0.958 \cite{segalMachineLearningAlgorithm2020}. P. Ventrella et al. (2021) evaluated the progression of CKD using data from Vimercate Hospital's Electronic Medical Records (EMR). They applied various machine learning algorithms, including ExtraTrees classifier, Decision Trees, Random Forest, Neural Networks, and Logistic Regression. The ExtraTrees classifier showed high accuracy in predicting the need for dialysis in CKD patients, with accuracies of 0.94 for two classes, 0.91 for three classes, and 0.87 for four classes  \cite{ventrellaSupervisedMachineLearning2021}. K.H. Lee et al. (2022) focused on predicting ESRD risk among sepsis survivors with CKD using algorithms like Random Forest, Extra Trees, XGBoost, LGBM, and Gradient Boosting Decision Tree (GBDT). The GBDT algorithm achieved the highest area under the receiver operating characteristic curve (AUC) of 0.879. Their study initially identified 112,628 CKD patients using ICD-9 and ICD-10 codes and, after applying exclusion criteria, enrolled 11,661 eligible sepsis survivors with a history of CKD \cite {leeArtificialIntelligenceRisk2022}.

Lastly, Murugesan G et al. (2022) integrated fuzzy logic with neural network capabilities through the ANFIS algorithm to analyze patient data related to various CKD risk factors. Their model achieved an accuracy of 93.75\%, with high sensitivity, specificity, and precision rates  \cite{g.FuzzyLogicBasedSystems2022}.

\subsubsection{Dementia and Alzheimer's Prediction}

The studies present significant advancements in predicting dementia and Alzheimer's disease using various machine learning models and datasets. Miled et al. (2020) employed a Random Forest (RF) model to predict dementia using routine care electronic medical records (EMR) data, achieving a generalizable model with 77.43\% accuracy, 76.01\% sensitivity, and 74.16\% specificity. Their dataset included both structured data like prescriptions and diagnoses, as well as unstructured data such as medical notes, encompassing features like age, gender, and race \cite{benmiledPredictingDementiaRoutine2020}.  El-Sappagh et al.(2021) developed a two-layered model using RF classifiers, Decision Trees, and Fuzzy Rule-Based Systems to predict cognitive states and Alzheimer's progression. They achieved cross-validation accuracies of 93.95\% and 87.08\% in the first and second layers, respectively, using data from the Alzheimer's Disease Neuroimaging Initiative (ADNI). Their model emphasized interpretability and trustworthiness with the SHAP framework  \cite{el-sappaghMultilayerMultimodalDetection2021}. The study by Yingjie Li et al. (2020) employs Principal Analysis through Conditional Expectation (PACE) to extract functional principal component scores from sparse and irregularly spaced measurements. This methodology achieved an overall sensitivity of around 80\%, specificity above 70\%, accuracy around 75\%, and AUC above 80\%. Notably, the entorhinal cortex (EC) was identified as the best predictor with AUCs above 0.83 for both 1-year and 2-year advanced predictions, utilizing data from the ADNIMERGE file, which includes MRI data and corresponding clinical information for subjects with mild cognitive impairment (MCI) \cite{liEarlyPredictionAlzheimer2020}.

Further advancing the methodology, Chatterjee et al. (2022) utilized a combination of Support Vector Machine (SVM), K-Nearest Neighbor (KNN), Logistic Regression, and Naive Bayes. Their approach achieved the highest accuracy of 96.43\% in classifying Alzheimer's Disease (AD). The study was based on the OASIS project data, including longitudinal and cross-sectional brain MRI data. This dataset consisted of 150 patients aged 18 to 96 years, with 72 patients categorized as having no dementia and 64 patients classified as having dementia at their first visits \cite{chatterjeeVotingEnsembleApproach2022}. 
Khan et al. (2022) integrated multiple models such as K-Nearest Neighbors (KNN), Naive Bayes (NB), Gradient Boosting (GB), XGBoost (XGB), and Support Vector Machine (SVM) with various kernels. They proposed an ensemble model combining XGB, Decision Tree (DT), and SVM, which outperformed all other models with an efficiency of 95.75\%. Their dataset comprised 2127 MRI (T1 and T2 weighted) images from the ADNI website, categorized into three classes: Alzheimer's Disease (AD), Mild Cognitive Impairment (MCI), and Control Normal (CN)  \cite{khanEnsembleModelDiagnostic2022}. Jahan et al. (2023) compared nine popular machine learning models for Alzheimer's prediction, demonstrating high accuracy scores ranging from 94.51\% to 98.94\% using the OASIS-3 dataset, which includes extensive longitudinal neuroimaging, cognitive, clinical, and biomarker data. Additionally, they innovated with the use of wearable sensor bands for continuous data collection, processed through a mist layer for real-time analysis \cite{jahanExplainableAIbasedAlzheimer2023}. 

\subsubsection{COPD, Lung, Liver, and Pancreatic Disease Prediction}

M. Moll et al. (2020) developed a Machine Learning Mortality Prediction (MLMP) model for chronic obstructive pulmonary disease (COPD) using Cox regression and random survival forest techniques. The MLMP model surpassed existing models like BODE and ADO in predicting all-cause mortality across two large COPD cohorts. It was developed using data from 2,632 participants in the COPDGene Study and validated on 1,268 participants from the ECLIPSE Study, achieving a C index of at least 0.7 \cite{mollMachineLearningPrediction2020}. In the realm of liver disease prediction, S. Rajesh et al. (2020) employed five ML algorithms—KNN, Naïve Bayes, decision tree, random forest, and SVM—to predict Hepatocellular Carcinoma (HCC). The study used two datasets, HCC STo from the UCI Machine Learning Repository and a Kaggle dataset, comprising 49 attributes. Random forest and KNN achieved the highest accuracies of 79.46\% and 79.04\%, respectively, with random forest reaching 80.64\% without cross-validation  \cite{rajeshHepatocellularCarcinomaHCC2020}. For lung disease prediction, S. Abbas et al. (2023) implemented a Fused Weighted Federated Deep Extreme Machine Learning (FDEML) model to predict lung cancer. This model integrated federated learning and deep extreme machine learning techniques, achieving a high accuracy of 96.30\%. The study addressed data fragmentation and privacy concerns by employing federated learning, allowing collaborative model development without sharing sensitive patient data. The dataset comprised 309 cases from IoMT sensors, supplemented by 231 records, highlighting the significance of IoMT-based data in enhancing prediction accuracy \cite{abbasFusedWeightedFederated2023}.   

In the realm of liver disease prediction, Barus et al. (2022) focused on employing supervised machine learning models, specifically Support Vector Machine (SVM) and Logistic Regression (LR). They used the Indian Liver Patient Dataset (ILPD), which contains 583 patient records with 11 features each. In their preprocessing stage, they applied Principal Component Analysis (PCA) and Synthetic Minority Over-sampling Technique (SMOTE) to evaluate their impact on the performance of the machine learning models. Initially, without PCA and SMOTE, LR achieved an accuracy of 70\%, while SVM reached 88\%. After incorporating PCA and SMOTE, the accuracy of LR decreased to 64\%, and SVM’s accuracy slightly dropped to 87\%. This indicates that the preprocessing techniques did not enhance and even slightly reduced the performance of the models in this context \cite{barusLiverDiseasePrediction2022}. Md et al. (2023) adopted a different approach by leveraging various ensemble learning algorithms to improve the accuracy of liver disease prediction using the same ILPD dataset. They explored models such as Gradient Boosting, XGBoost, Bagging, Random Forest, Extra Tree, and Stacking. The results were promising, with Gradient Boosting achieving a testing accuracy of 91.82\% and Stacking ensemble attaining 86.06\%. These findings highlight the superiority of ensemble learning methods in enhancing predictive accuracy, outperforming the single models used in other studies \cite{mdEnhancedPreprocessingApproach2023}. Likewise,
Kumar et al. (2023) conducted a comprehensive evaluation using a broad spectrum of machine learning models on the ILPD dataset. Their study included Decision Tree (DT), K-Nearest Neighbors (KNN), Multilayer Perceptron (MLP), AdaBoost (AB), Random Forest (RF), Gradient Boosting (GB), XGBoost (XGB), Logistic Regression (LR), Naive Bayes (NB), Extra Tree (ET), LightGBM (LGBM), and SVM. Among these, the Decision Tree model exhibited robust performance, with an accuracy of 86.67\%, precision of 0.87, recall of 0.87, and F1-Score of 0.86. This suggests that DT is a viable and effective model for liver disease prediction, offering balanced and reliable metrics across various performance indicators \cite{kumarLiverDiseasePrediction2023}.

In addition, Warda M. Shaban (2023) in Liver Patients Detection Strategy (LPDS) employs machine learning classifiers such as SVM, KNN, Naive Bayes (NB), Decision Tree (DT), and Random Forest (RF) on a training set to classify patients and determine infection status through data preprocessing, feature selection, and classification phases. The LPDS model's performance metrics include an accuracy of 89.5\%, sensitivity of 91.2\%, specificity of 87.3\%, F1 score of 0.89, and AUC of 0.92. Additionally, the KNN classifier obtained the highest classification accuracy of 99.1\% on the test dataset. The dataset used is the Kaggle Liver dataset with a total of 441 records \cite{wardam.shabanEarlyDiagnosisLiver2023}. Similarly, Ruhul Amin et al. (2023) in their study,  combines Factor Analysis (FA) with Linear Discriminant Analysis (LDA) for feature selection and classification. The model uses machine learning techniques such as Random Forest (RF), K-Nearest Neighbors (KNN), Logistic Regression (LR), Multi-Layer Perceptron (MLP), Support Vector Machine (SVM), and Boosted Regression Trees (BRT) to predict and diagnose liver diseases. The system achieves an accuracy of 88.10\%, precision of 85.33\%, recall of 92.30\%, F1 score of 88.68\%, and an AUC score of 88.20\%. The study's results outperform existing studies by 0.10-18.5\%. Further exploration and fine-tuning of model parameters are recommended to improve the AUC score. The dataset used is the Indian Liver Patient Dataset (ILPD) from the UCI machine learning repository, consisting of 583 observations with ten liver function-related features and one target output \cite{aminPredictionChronicLiver2023}.

In the realm of pancreatic disease, Li et al. (2018) developed a Computer-Aided Detection (CAD) model for pancreatic cancer that involves three main steps: pancreas segmentation, feature extraction and selection, and classifier design. The segmentation process uses Simple Linear Iterative Clustering (SLIC) on CT pseudo-color images from the General Image Processing (GIP) method. Feature selection and combination are performed using Decision Tree-Principal Component Analysis (DT-PCA), and a hybrid model combining Hierarchical Feature Bagging-Support Vector Machine-Random Forest (HFB-SVM-RF) is used for classification. The model achieved an accuracy of 96.47\%, a sensitivity of 95.23\%, and a specificity of 97.51\%. It was evaluated using the National Institutes of Health (NIH) dataset, obtaining a mean Dice coefficient of 78.9\% and a Jaccard index of 65.4\%. The dataset comprised PET/CT data from 80 patients, each with approximately 1700 DICOM images, and the labels were confirmed by three nuclear medicine experts  \cite{liEffectiveComputerAided2018}. 
W. Muhammad et al. (2019) focused on predicting pancreatic cancer risk using an artificial neural network (ANN) model. This study utilized data from the NHIS and PLCO trials, incorporating 18 personal health features to develop and train the model achieving high sensitivity and specificity, with an AUC of 0.85 for the test dataset and 0.86 for the training dataset  \cite{muhammadPancreaticCancerPrediction2019}.

In a related application, Y. Zhou et al. (2022) developed predictive models for the severity of acute pancreatitis (AP) using five ML algorithms: logistic regression (LR), support vector machine (SVM), decision tree (DT), random forest (RF), and XGBoost. Data preprocessing included Minmax scaling, one-hot encoding, and handling missing values with the MissForest package. XGBoost showed superior performance with an AUC of 0.906, high accuracy, sensitivity, specificity, and an F1 score of 0.764 \cite{zhouPredictionSeverityAcute2022}. B. Kui et al. (2022) also tackled the prediction of acute pancreatitis severity using ML algorithms such as decision tree, random forest, logistic regression, SVM, CatBoost, and XGBoost. Addressing missing data and imbalanced classes with KNNImputer and SMOTE, XGBoost achieved the best performance with an AUC of 0.81 and an accuracy of 89.1\%. This study involved an international cohort of 1184 patients for model development and 3543 for validation, utilizing data from multiple centers to enhance model generalizability. Additionally, the study developed an easy-to-use web application to improve model accessibility and usability \cite{kuiEASYAPPArtificial2022}.

\subsubsection{Multiple Diseases Prediction}

N.K. Sbehara et al. (2015) conducted a study on the classification of thyroid, hepatitis, and heart diseases using Multilayer Perceptron (MLP) models enhanced by Bird Mating Optimization (BMO) and Firefly Algorithm (FFA). Their research demonstrated that MLP-BMO generally outperformed MLP-FFA, especially for heart and liver diseases, as evidenced by Mean Squared Error (MSE) comparisons. Specifically, the MSE for liver diseases showed MLP-FFA with a range from 2.8892 to 4.1277 (average 2.9456, standard deviation 0.4425), whereas MLP-BMO had a range from 2.8002 to 4.0714 (average 2.9802, standard deviation 0.3534) \cite{beheraBirdMatingOptimization2015}.

Akkem Yaganteeswarudu (2020) conducted a study on multi-disease prediction using machine learning, TensorFlow, and Flask API, with a thorough parameter analysis to optimize disease effect detection. The study utilized Python pickling for saving and loading model behavior and incorporated various machine learning and deep learning techniques, including logistic regression, Naïve Bayes, SVM, decision tree, and random forest algorithms. Notably, logistic regression achieved a 92\% accuracy rate for diabetes analysis, random forest yielded 95\% accuracy for heart disease classification, and SVM reached 96\% accuracy for cancer detection. Additionally, TensorFlow convolutional neural networks were employed for diabetic retinopathy analysis, achieving 91\% accuracy. The study's model demonstrated impressive performance with high accuracy rates across different diseases. The datasets used included the Pima Indian Diabetes Dataset for diabetes analysis, over 150 GB of image data from the UCI machine learning repository for diabetic retinopathy prediction, heart disease patient data from Cleveland, Hungarian, and Switzerland, and the Breast Cancer Wisconsin (Diagnostic) Data Set for cancer prediction and live data sets from corresponding hospitals \cite{yaganteeswaruduMultiDiseasePrediction2020}.

P. Singh et al. (2021) presented a diagnostic model for heart disease and multi-disease conditions that integrates Particle Swarm Optimization (PSO) and beetle foraging behavior for optimization within a five-layer structure governed by fuzzy rules. Evaluated via MATLAB, the BSO-ANFIS model achieved impressive results, with an accuracy of 99.1\% and precision of 99.37\% for heart disease classification, and an accuracy of 96.08\% and precision of 98.63\% for multi-disease classification. The study suggests potential improvements through further accuracy investigation and the integration of diverse data types such as images, audio, or video. The heart disease dataset was sourced from Kaggle, while the multi-disease dataset was from USA healthcare services \cite{singhMultidiseaseBigData2021}.

J. Rash et al. (2022) focused on predicting five chronic diseases—diabetes, breast cancer, hepatitis, heart disease, and kidney disease—using an Artificial Neural Network (ANN) optimized by Particle Swarm Optimization (PSO) achieving a peak accuracy of 99.67\%, surpassing other benchmark methods and demonstrating faster processing times compared to Random Forest (RF), deep learning, and Support Vector Machine (SVM). Specifically, ANN-PSO outperformed traditional methods like Logistic Regression (LR: 68.97\% accuracy), SVM (98.23\% accuracy), K-Nearest Neighbors (KNN: 94.69\% accuracy), Naive Bayes (NB: 97.35\% accuracy), and Decision Trees (DT: 95.58\% accuracy). The datasets utilized were obtained from online sources like Kaggle, Dataworld, Github, and the UCI machine learning repository, ensuring a diverse and comprehensive data foundation for the study \cite{rashidAugmentedArtificialIntelligence2022}.

Table \ref{tab:ML_publications} summarizes all the thirty-four studies that apply various ML techniques such as RF, DT, LR, XGBoost, SVM, KNN, NB, LGBM, GBDT, ANN, etc., to predict and analyze chronic diseases. Each study is categorized by disease type and ML model used, with reported performance metrics such as Accuracy, Sensitivity, Specificity, ROC AUC, and F1-score. Dataset descriptions encompass diverse sources from medical imaging to clinical records. Notably, integration with Internet of Medical Things (IoMT) data and stringent data privacy measures are highlighted across studies to protect.

\subsection{DL Techniques Deployed for Terminal Disease Prediction}
The integration of deep learning algorithms with the Internet of Medical Things (IoMT) transforms healthcare diagnostics and disease management, especially for chronic and severe conditions like cardiovascular diseases, chronic kidney disease, Alzheimer's, stroke, COPD, and liver disease. These technologies use algorithms such as Convolutional Neural Networks (CNNs), Enhanced Deep Learning Assisted Convolutional Neural Network (EDCNN), Deep Learning Modified Neural Networks (DLMNN), Adaptive Hybridized Deep Convolutional Neural Network (AHDCNN), hybrid fuzzy deep neural network (HFDNN), Bi-LSTM, and other hybrid models. CNNs, with their ability to analyze medical imaging data precisely, help healthcare providers detect subtle abnormalities more accurately. Innovations like EDCNN and DLMNN refine standard CNN architectures, enhancing their ability to interpret complex medical data patterns. AHDCNN and HFDNN models combine deep learning with fuzzy logic or adaptive learning mechanisms, allowing nuanced decision-making in diagnostics. The IoMT enables continuous monitoring and real-time analysis of patient data for early detection and timely intervention in chronic diseases.

In the following subsections, we will provide a detailed analysis of each family of chronic and terminal diseases prediction using deep learning models

\subsection{CVD (Heart Disease) Prediction}
The studies in this subsection underscore the transformative potential of deep learning and IoT in CVD-Heart Disease prediction and diagnosis. Using algorithms like CNNs, DLMNN, Bi-LSTM, and hybrid models, they show significant improvements in accuracy, specificity, and sensitivity. IoMT enables continuous monitoring and real-time analysis, crucial for early detection. Emphasis on data security through encryption enhances patient privacy and trust. Y. Pan et al. (2020) introduced an Enhanced Deep Learning Assisted Convolutional Neural Network (EDCNN) for precise heart disease prediction and diagnosis, achieving an impressive accuracy of 97.51\%, with sensitivity and specificity scores of 93.51\% and 94.9\%, respectively. Utilizing clinical data from the UCI repository, including blood pressure, heart rate, and ECG signals, and proposed a Remote Patient Monitoring (RPM) platform using IoMT to collect and transmit health parameters, though privacy considerations were not explicitly detailed  \cite{panEnhancedDeepLearning2020}. In 2020, S. S. Sarmah presented a Deep Learning Modified Neural Network (DLMNN) optimized with the Cuttlefish Optimization Algorithm (CFOA) for heart disease prediction using the Hungarian Heart Disease dataset, achieving an accuracy of 92\%, outperformed traditional Artificial Neural Networks (ANN) and demonstrated superior sensitivity, specificity, and f-measure. The study emphasizes using a network of body sensors to collect and analyze vital signs such as breathing rates, blood pressure, pulse, and body temperature. This collected data is then transmitted to a healthcare application where it is stored and analyzed for medical purposes. Secure data transmission via PDH-AES encryption and compression with the Modified Huffman Algorithm (MHA), highlighting the importance of data security and efficient storage in healthcare \cite{sarmahEfficientIoTBasedPatient2020}.

S. Hussain et al. (2021) proposed a novel 1D Convolutional Neural Network (CNN) architecture, achieving over 97\% training accuracy and 96\% test accuracy on the Cleveland heart disease dataset.Recommendations included integrating wearable sensors and expanding parameters to enhance the model's robustness and applicability \cite{hussainNovelDeepLearning2021}. A. Kumar et al. (2022) explored various models, including Naive Bayes, SVM, k-NN, CNN, and ANN, for early heart attack prediction. The CNN model outperformed others with an accuracy of 98\%, precision of 97\%, and an average F-score of 98\%. Using the UCI dataset, suggesting higher performance of deep learning models \cite{kumarECGBasedEarly2022}. Similarly, S. M. Nagarajan et al. (2022) developed a model combining Decision Trees (DT), SVM, Random Forest (RF), and Naive Bayes (NB) into an ensemble classifier with a Deep Convolutional Neural Network (DCNN). The model achieved a classification accuracy of 94\% with original features and 95.34\% with extracted features across ten different medical datasets,suggesting higher performance of deep learning models \cite{nagarajanInnovativeFeatureSelection2022}.  S. Manimurugan et al. (2022) proposed a two-stage classification model using hybrid linear discriminant analysis with modified ant lion optimization for sensor data and a hybrid Faster R-CNN with SE-ResNet-101 for echocardiogram images. The model achieved up to 99.15\% accuracy for image classification utilizing the Cleveland dataset.The study highlighted IoMT's role in securely collecting and transmitting sensor data to the cloud. Data privacy was maintained by storing medical data in a secure cloud database \cite{manimuruganTwoStageClassificationModel2022}. A. A. Nancy et al. (2022) combined a fuzzy system with a Bi-LSTM model for heart disease risk prediction, achieving performance metrics ranging from 98.03\% to 98.90\%. Utilizing augmented Cleveland and Hungarian datasets up to 100,000 records, the study stressed the need for robust encryption, access controls, and ethical guidelines to ensure patient data privacy, highlighting the pivotal role of IoMT in remote patient monitoring and real-time data transmission to the cloud \cite{nancyIoTCloudBasedSmartHealthcare2022}. L. Kumar et al. (2023) utilized Convolutional Neural Networks (CNNs) trained on ECG signals and patient clinical data, achieving a 98.5\% accuracy rate. The research dataset for heart disease prediction includes data from 2015 to 2020, covering a total of 316,980 individuals. This includes 2,976 individuals with heart failure, 18,203 individuals closely connected to heart failure patients, and 295,801 healthy individuals.\cite{kumarDeepLearningBased2023}. S. Deepa et al. (2023) developed the ELHE model, a deep learning approach for cardiovascular disease prediction, achieving over 97\% accuracy. The study employed UCI laboratory datasets on heart disease patients, utilizing two types: one comprehensive dataset and another with select attributes \cite{sExperimentalEvaluationArtificial2023}.

Lastly, M. Munsif et al. (2024) proposed a hybrid classification model combining Genetic Algorithm (GA) for feature selection, SVM for initial classification, and CNN for final classification. The model achieved accuracies of 98\% on the UCI dataset, 97\% on the Z-Alizadeh Sani dataset, and 86\% on the Cardiovascular Disease dataset. The study recommended validating the GA-SVM-CNN model on diverse datasets, ensuring data privacy with encryption and access controls, optimizing computational costs, addressing class imbalance, and enhancing model interpretability for healthcare insights \cite{munsifEfficientHybridClassification2024}.

\subsubsection{CKD/ESRD}
Several notable studies have investigated the use of advanced methodologies for kidney disease prediction and diagnosis. G. CHEN et al. (2020) utilized an Adaptive Hybridized Deep Convolutional Neural Network (AHDCNN) integrated with the Internet of Medical Things (IoMT) for early detection of chronic kidney disease. This approach achieved high precision and recall ratios, with an F1-score indicating balanced performance, and demonstrated 80\% successful segmentation of kidneys using data from the DeepLesion database. The IoMT implementation allowed for remote tracking and monitoring of patients, transforming medical supervision into telemedicine  \cite{chenPredictionChronicKidney2020}. Similarly, M. U. Nasir et al. (2022) developed a deep learning framework employing transfer learning techniques and optimization algorithms (SGDM, ADAM, RMSPROP) combined with IoMT and blockchain for early kidney cancer prediction. Their model attained a training accuracy of 99.8\% and prediction accuracy of 99.20\%, leveraging data augmentation and transfer learning to improve prediction accuracy and reduce misclassification rates. The dataset included 3300 augmented samples across three grades, with IoMT and blockchain enhancing data quality and security \cite{nasirKidneyCancerPrediction2022}.

In another study, K. Kumar et al. (2023) focused on a hybrid fuzzy deep neural network (FDNN) for early-stage prediction of chronic kidney disease, integrating fuzzy inference systems with neural networks. This Hybrid Fuzzy Neural Network (HFNN) demonstrated superior accuracy (99.23\%) compared to existing methods (97.46\%), particularly in managing CKD comorbidities and focusing on hypertension values. The dataset consisted of 5617 records from Changhua Christian Hospital, utilizing cameras as sensor for patient images for disease detection \cite{kumarDeepLearningApproach2023}. Meanwhile, D. M. Alsekait et al. (2023) proposed a two-level prediction model using pretrained models (RNN, GRU, LSTM) in Level 1 and a metalearner (SVM) in Level 2 for stacking ensemble techniques. This model used the Chronic Kidney Disease dataset from the UCI machine learning repository, containing 400 cases with various features, and employed feature selection and preprocessing methods like chi-square test, recursive feature elimination, and handling of missing values to improve the prediction model. The RNN Layer2 achieved the highest performance metrics, with the proposed model's Layer2 also demonstrating high performance \cite{alsekaitComprehensiveChronicKidney2023}.

\subsubsection{Alzheimer’s and stroke Prediction}
In predicting diseases such as Alzheimer’s and stroke, models like LSTM RNNs and deep neural networks, employing techniques such as dropout, regularization, and optimization algorithms like Adam have achieved high accuracies in predicting  the diseases, utilizing datasets from national health surveys and ensuring real-time data processing and enhances the reliability of medical prognostics. T. Wang et al. ( 2018) in  Alzheimer’s disease prediction utilizing dataset from the National Alzheimer's Coordinating Center (NACC), involving 5432 patients in a two-layer Long Short-Term Memory (LSTM) recurrent neural network (RNN) with 100 hidden units per layer was utilized. Techniques included learning rate decay, moving average decay, L2 regularization, and the Adam Optimizer. Data preprocessing involved mean, median, and mode imputation, normalization, and encoding. Key features were Clinical Dementia Rating (CDR), Geriatric Depression Scale (GDS), and Functional Activities Questionnaire (FAQ), achieving  superior performance metrics of Accuracy: 0.9906 ± 0.0043; PPIA: 0.9894 ± 0.0074; SPIA: 0.9912 ± 0.0039 \cite{wangPredictiveModelingProgression2018}. Dua et al. (2020) explored the use of Convolutional Neural Networks (CNN), Recurrent Neural Networks (RNN), and Long Short-Term Memory (LSTM) models. Each model underwent Bagging to reduce variance, and the Bagged models were then combined using a weighted average ensemble technique. This approach led to individual model accuracies of 89.75\%, with the ensemble model achieving an impressive accuracy of 92.22\%. The datasets used were OASIS-1, containing cross-sectional MRI data of 416 subjects, and OASIS-2, which includes longitudinal MRI data of 150 subjects \cite{duaCNNRNNLSTM2020}. Similarly, Jungyoon Kim  and Jihye Lim (2021) used dataset from from the Korea National Health and Nutrition Examination Survey (KNHANES), with 7031 participants aged over 65 to train a  Deep Neural Network (DNN) model with four hidden layers (30 neurons each) for dementia prediction using ReLU activation for hidden layers and a sigmoid function for the output. Techniques included 0.4 dropout probability, Adam optimization, binary cross-entropy, and a learning rate of 0.001. Scaled Principal Component Analysis (PCA) was used for data preprocessing, achieving Performance metrics of AUC: 85.5\%, Recall: 68.6\%, Specificity: 82.1\%, Precision: 5.\%, Accuracy: 81.9\% \cite{kimDeepNeuralNetworkBased2021}.

Building on these techniques, Y-A Choi et al. (2021) utilized various deep learning models (LSTM, Bidirectional LSTM, CNN-LSTM, CNN-Bidirectional LSTM) for stroke  prediction using on real-time EEG sensor data validated using five-fold cross-validation.The CNN-Bidirectional LSTM model achieved the highest accuracy (94.0\%), precision (94.6\%), and F1-score (94.1\%) \cite{choiDeepLearningBasedStroke2021}. Chui et al. (2022) adopted a novel approach by enhancing CNN models through transfer learning to incorporate domain knowledge from diverse datasets, and using a Generative Adversarial Network (GAN) to generate additional data for minority classes. This combination improved the detection model's accuracy by 2.85–3.88\%, 2.43–2.66\%, and 1.8–40.1\% in different scenarios. They utilized the OASIS-1, OASIS-2, and OASIS-3 datasets, which included 416, 150, and 1098 participants respectively, and covered a wide range of classes from Normal to Moderate AD \cite{chuiMRIScansBasedAlzheimer2022}. Odusanmi et al. (2023) employed a multimodal fusion-based approach using discrete wavelet transform (DWT) to analyze data, optimized through transfer learning with a pre-trained neural network (VGG16). The final fused image was reconstructed using inverse discrete wavelet transform (IDWT). This method achieved an accuracy of 81.25\% for AD/EMCI and AD/LMCI in MRI test data, and 93.75\% for AD/EMCI and AD/LMCI in PET test data. Their study utilized sMRI and FDG-PET images from the ADNI database, focusing on participants aged 60 to 70 years from the ADNI2 baseline cohort, including approximately 50 early EMCI and 50 LMCI participants \cite{odusamiPixelLevelFusionApproach2023}.

On Stroke Prediction,  Elbagoury et al. (2023 present a GMDH-type polynomial network with linear activation functions, achieving a training error of 0.001 and operating within a mobile AI smart hospital platform. The model, leveraging feature scaling techniques for data preprocessing and GMDP deep learning models for feature extraction from EMG signals, demonstrates high accuracy (96.85\%) in predicting EMG signals. The study utilizes datasets including the EMG Lower Limb Dataset and mHealth Dataset, integrating IoMT technology for real-time analysis and prediction of heart and stroke conditions, thereby showcasing advanced AI applications in healthcare \cite{ elbagouryHybridStackedCNN2023}.

\subsubsection{COPD, Lung, Liver and Pancreatic Disease Prediction}
In a study by C.T. Wu et al. (2021), researchers developed a deep neural network (DNN) to predict acute exacerbations of chronic obstructive pulmonary disorder (AECOPD), achieving over 90\% accuracy. The DNN used 45 features including physiological, environmental, and lifestyle data from training dataset consisted of 5600 data points. Evaluation metrics on a test dataset showed high performance with an F1 score of 0.9323, precision of 0.9393, specificity of 0.9253, sensitivity of 0.9452, AUROC of 0.9699, and accuracy of 0.9357. (IoMT) devices such as wearables and home sensors are used for automated data input. To ensure privacy, the system used HTTPS and encryption protocols, restricting data access to verified medical providers, thus maintaining confidentiality and integrity \cite{wuAcuteExacerbationChronic2021}.  Conversely, the study by S. L. Jany Shabu et Al. (2023) employs a deep neural network with residual connections and an adaptive neuro-fuzzy inference system (ANFIS) for lung cancer detection and prediction, utilizing the Internet of Medical Things (IoMT). The deep neural network includes padding, pooling, dropout, convolution layers, 16 residual units, and two fully connected layers to enhance efficiency and performance. ANFIS combines fuzzy systems' learning skills with neural networks' cognitive abilities, improving prediction accuracy and resource utilization. The FES-LC model achieves an accuracy of 94.2\% in lung cancer detection and prediction, with a mean absolute error ranging from 52.3\% to 62.12\% across iterations. It recommends using the FES-LC alongside simulations and the MNP approach for accurate lung cancer detection. The dataset used is a privately sourced balanced lung tumor dataset \cite{shabuImprovedAdaptiveNeurofuzzy2023}.

Similarly,  the study by T.A. Khan et al. (2023) focused on Google Net Transfer Learning (TL) model applied in private edge clouds for lung cancer disease prediction, trained on a dataset of 25,000 images of lung and colon cancer tissues. The performance of the model was measured using accuracy, misclassification rate, precision, recall, specificity, and F1-score, achieving a high accuracy of 98.8\%, thereby outperforming previous methodologies. The TL approach was enhanced through data collected from IoMT devices in hospitals, aiming to improve prediction accuracy while preserving patient privacy in smart healthcare 5.0 environments. Additionally, the model incorporated a Federated Learning (FL) approach to ensure data privacy, where data was collected from multiple hospitals and transferred to a central database securely using IoMT devices \cite{khanSecureIoMTDisease2023}. Hefu Xie et al. (2024) proposed model integrates a fully connected neural network for basic patient features and a convolutional neural network for capturing temporal patterns in patient data. The combination allows automatic learning of complex patterns and abstract features in highly dynamic medical data associated with acute liver failure (ALF). This deep learning model integrates fully connected and convolutional neural networks, outperforming traditional machine learning methods such as Logistic Regression, K-Nearest Neighbors, SVM, Decision Tree, Random Forest, XGBoost, and Gradient Boosting in terms of accuracy and generalization, achieving 94.8\% accuracy. The dataset includes diverse patient indicators like age, gender, weight, obesity status, blood pressure, cholesterol levels, hepatitis status, and family history of hepatitis \cite{xieDeepLearningApproach2024}.

For pancreatic disease, Asaturyan et al. (2019) employed a methodology using 3D techniques on MRI and CT scans for pancreatic segmentation. Their approach involves three main steps: enhancing the pancreas region, applying 3D segmentation using max-flow and edge detection techniques, and removing non-pancreatic contours. The model achieved a mean Dice Similarity Coefficient (DSC) of 79.3 ± 4.4\%. They evaluated their method on two MRI datasets containing 216 and 132 image volumes, achieving mean DSCs of 79.6 ± 5.7\% and 81.6 ± 5.1\%, respectively. Additionally, they used a dataset containing 82 CT image volumes for their analysis \cite{asaturyanMorphologicalMultilevelGeometrical2019}. Hu et al. (2021) introduced the DSD-ASPP-Net architecture, which integrates DenseASPP with saliency-aware modules for automatic pancreas segmentation. This advanced architecture achieved an average Dice-Sørensen Coefficient (DSC) of 85.49 ± 4.77\%, surpassing the performance of previous methods. They evaluated their approach on the public NIH pancreas dataset and a local hospital dataset, demonstrating its robustness and effectiveness in clinical settings  \cite{huAutomaticPancreasSegmentation2021}.

\subsubsection{Multiple Disease Prediction}
Just as machine learning models can be trained on datasets to predict multiple diseases, deep learning models can perform even better due to their advanced capabilities in handling complex data, extracting intricate features, and improving accuracy in disease prediction and diagnosis.Researchers have developed sophisticated models combining deep learning techniques, such as convolutional neural networks and ensemble learning, with IoT frameworks to improve accuracy and efficiencyleveraging synthetic data generation, optimization algorithms, and secure data handling to predict various conditions, including heart disease, diabetes, cancer, and Alzheimer's,etc.  In this study of S. Thandapani et al. (2023), a pandemic support system was developed utilizing various deep convolutional network architectures, including ResNet (50, 100, 101) and VGG (16, 19) to train datasets comprised CT and X-ray images from diverse sources like Kaggle, GitHub, and hospitals, enhanced by generating synthetic data using the Keras image data generator. To address individual model bias, a major voting classifier was employed, aggregating the results from multiple models to improve classification accuracy. ResNet 101 and VGG 19 achieved the highest accuracies for CT and X-ray images, respectively.  The system's IoMT implementation integrated sensors attached to patients, allowing real-time monitoring via the internet and cloud technology \cite{thandapaniIoMTDeepCNN2023}. A. S. Prakaash et al. (2022) presented a multi-disease prediction model leveraging ensemble learning with weighted RBM features, combining classifiers such as DNN, SVM, and RNN. The ASR-CHIO algorithm was employed to optimize parameters, significantly enhancing prediction accuracy. The model outperformed traditional algorithms, achieving high F1-scores, MCC, specificity, and NPV. Datasets from various sources, including Kaggle and UCI, covering diseases like COVID-19, EEG Eye State, Epileptic Seizure, Stroke Prediction, Heart, and Diabetic Retinopathy, were utilized. The advanced optimization techniques employed ensured the model's robustness and efficiency in predicting multiple diseases accurately \cite{prakaashDesignDevelopmentModified2022}.

N. Nigar et al. (2023) focused on diagnosing chronic diseases such as COVID-19, pneumonia, diabetes, heart disease, brain tumors, and Alzheimer's using five pretrained deep CNN models: VGG16, VGG19, ResNet, DenseNet, and Inception-v3. These models were trained with hyperparameter optimization and the XGBoost algorithm, with data preprocessing and augmentation enhancing their performance. Evaluations based on accuracy, precision, recall, and F1 score demonstrated varied success rates across different diseases. The study utilized real-time datasets from Kaggle and other sources and implemented an IoMT framework to connect medical devices and sensors, employing enhanced Elliptic Curve Cryptography (ECC) to ensure patient data privacy \cite{nigarIoMTMeetsMachine2023}. M. K. Chowdary et al. (2023) utilized a combination of machine learning algorithms (Random Forest, Decision Tree, Gradient Boosting, Logistic Regression, XGBoost, SVM, KNN) and deep learning (VGG19 CNN) to predict diseases like heart disease, diabetes, breast cancer, liver disease, kidney disease, malaria, and pneumonia. The model demonstrated high accuracy for various diseases, with SVM, Random Forest, and VGG19 showing exceptional performance metrics. Datasets from sources like Kaggle, NIH, and medical centers were employed, highlighting the algorithms' potential in predicting multiple diseases with high precision and reliability. Techniques like ROC analysis were used to evaluate performance, showcasing the robustness of the model \cite{chowdaryMultipleDiseasePrediction2023}. 

Anusha Ampavathi and T. Vijaya Saradhi (2021) developed a hybrid deep learning model using the JA-MVO-RNN + DBN algorithm, implemented in MATLAB, for predicting multiple diseases, including diabetes, hepatitis, lung cancer, liver tumor, Alzheimer's, Parkinson's, and heart disease. The predictive models demonstrate high performance with the following highest figures: Diabetes: Precision 0.97753, MCC 0.93528; Hepatitis: Precision 0.98684, MCC 0.84824; Lung cancer: Precision 0.97297, MCC 0.81523; Liver tumor: Precision 0.97753, MCC 0.93528; Alzheimer’s disease: Precision 1.0, MCC 0.94403 \cite{ampavathiMultiDiseasepredictionFramework2021}. 

R. Daid et al. (2021) employed a deep learning-based multi-layer convolution neural network using TensorFlow and Keras for the early detection of diseases such as heart disease, stroke, lung cancer, colorectal cancer, and depression. The methodology included data preprocessing, normalization, and augmentation. The model achieved high accuracy and precision across different training and test splits. The dataset comprised chronic symptoms, illness severity, diagnosis information, and environmental factors from 600,000 patients, providing a comprehensive basis for robust predictive modeling \cite{daidEffectiveMechanismEarly2021}.  

Table \ref{tab:DL_publications} summarizes thirty-three studies that apply various deep learning (DL) techniques, including VGG19 CNN, NN, RNN, DBN, LSTM, GoogleNet, VGG16 CNN, ResNet, and Inception, to predict and analyze chronic diseases. Each study is categorized by disease type and ML model used, with reported performance metrics such as Accuracy, Sensitivity, Specificity, ROC AUC, and F1-score. Dataset descriptions encompass diverse sources from medical imaging to clinical records. Notably, integration with Internet of Medical Things (IoMT) data and stringent data privacy measures are highlighted across studies to protect patient information.

\subsection{Comparative Analysis of Technologies Used Across Different Disease Types}
Significant advancements in cardiovascular, chronic kidney, and Alzheimer's predictions underscore the effectiveness of ML and DL models. Yet, diseases like COPD, dementia, stroke, and certain cancers remain underexplored, signaling critical areas for future research, especially in understanding disease comorbidities.

\subsubsection{Heart Disease Prediction}

Heart disease remains a leading cause of global mortality, underscoring the critical importance of early detection and treatment. Machine learning (ML) models, such as XGBoost, and deep learning (DL) models, like EDCNN, have proven highly effective in predicting heart disease. XGBoost stands out for its exceptional accuracy and precision in cardiovascular studies. Its strength lies in its ability to handle large, complex datasets with numerous features, manage missing data effectively, and avoid overfitting—common challenges in medical data analysis.

On the other hand, EDCNN and Faster R-CNN combined with SE-ResNet-101 demonstrate high precision in image-based predictions. These deep learning frameworks excel at extracting intricate patterns from imaging data, which is vital in cardiology for detailed visualization. For instance, XGBoost models (A. Ogunpola et al., 2024) achieve impressive F1 scores of 98.71\%, highlighting their robustness in managing cardiovascular data complexities \cite{ogunpolaMachineLearningBasedPredictive2024}. Similarly, deep learning models such as HLDA-MALO, integrated with Faster R-CNN (S. Manimurugan et al., 2022), achieve even higher precision, with F1 scores reaching up to 99.02\%, reflecting their advanced feature extraction capabilities \cite{manimuruganTwoStageClassificationModel2022}.
\subsubsection{Chronic and End-Stage Kidney Disease}
Kidney diseases, particularly chronic and end-stage renal disease, require continual monitoring and early prediction to prevent progression and manage treatments effectively. DL Models like Stacking Ensemble (involving RNN, LSTM, GRU, and SVM) excel due to their ability to learn temporal dependencies and complex patterns in patient data over time, which is crucial for monitoring kidney function and disease progression. Probabilistic Neural Networks (PNN) have shown promising results, particularly in early stages of kidney diseases. PNNs are adept at classifying complex patterns and are beneficial in settings where diagnosis is based on a clear set of measurable stages. Advanced DL ensembles (D. M. Alsekait et al., 2023) report near-perfect metrics (99.69\% across accuracy, precision, recall, F1 score), highlighting their potential in precise renal function analysis \cite{alsekaitComprehensiveChronicKidney2023}.

\subsubsection{Alzheimer’s, Dementia  and Stroke Prediction}

Alzheimer's disease and other forms of dementia pose significant challenges due to their progressive nature and the subtle onset of symptoms. The Random Forest (RF) algorithm has proven effective in predicting Alzheimer's, largely because of its ensemble approach, which adeptly handles diverse data types and is resilient to overfitting—a common issue with the heterogeneous and noisy datasets typical in Alzheimer's research. RF has demonstrated impressive accuracy in Alzheimer's prediction, achieving 98.81\% as reported by S. Jahan et al. (2023) \cite{jahanExplainableAIbasedAlzheimer2023}. However, models for dementia exhibit greater variability, highlighting the need for more targeted model development.

Deep Neural Networks (DNNs) combined with scaled Principal Component Analysis (PCA) offer a sophisticated approach to dementia prediction by leveraging dimensionality reduction to emphasize the most pertinent features. This method addresses the complex, multifactorial nature of dementia and has achieved an accuracy of 81.9\%, as detailed by Jungyoon Kim and Jihye Lim (2021) \cite{kimDeepNeuralNetworkBased2021}.

Furthermore, hybrid architectures like the Stacked CNN and Residual Feedback GMDH-LSTM models, used in stroke prediction, illustrate the potential of combining different model types to enhance predictive performance. These hybrid models demonstrate deep learning's ability to manage acute medical events with high accuracy, achieving 99\% accuracy according to Elbagoury et al. (2023) \cite{elbagouryHybridStackedCNN2023}.

\subsubsection{ COPD and Lung Cancer}

Both diseases benefit significantly from advanced imaging techniques, which are enhanced by DL models capable of detailed image analysis. Deep Neural Networks (DNN) show substantial efficacy in diagnosing COPD by analyzing complex patterns in pulmonary imaging data, which often includes subtle signs that precede clinical symptoms. IoMT-enabled GoogleNet with transfer learning offers a cutting-edge approach for lung cancer prediction, combining the strengths of deep learning with real-time data acquisition capabilities of IoMT devices.
Recent models demonstrate significant advancements, with DNNs achieving over 92\% accuracy in COPD prediction (C.T. Wu et al., 2021) \cite{wuAcuteExacerbationChronic2021} and hybrid models reaching up to 98.8\% in lung cancer diagnostics (T. A. Khan et al. 2023) \cite{khanSecureIoMTDisease2023}.

\subsubsection{Pancreatitis and Liver Disease Prediction} 

Acute conditions like pancreatitis and liver diseases require rapid and accurate prediction to enhance treatment outcomes.
XGBoost excels in feature extraction, helping identify the most relevant features from the dataset, a crucial capability for acute pancreatitis prediction as it enhances prediction accuracy. Additionally, XGBoost's scalable nature makes it efficient with large datasets, using gradient boosting to build trees sequentially, which is effective for handling complex datasets and missing values. Study by (Zhou et al.,2022)  \cite{zhouPredictionSeverityAcute2022} demonstrated XGBoost's high performance,  achieving an AUC of 0.906.

Li et al. (2018) and Xie et al. (2024) rely on the strength of  Hybrid models in combining advantages of the combining models.  Li et al. (2018) utilized a hybrid model that integrates Histogram of Oriented Gradients (HFB), Support Vector Machines (SVM), and Random Forest (RF) leveraging the strengths of each method and achieved accuracy of 96.47\%, sensibility of 95.23\% and specificity of 97.51\% \cite{liEffectiveComputerAided2018}. Hybrid deep learning models, which combine fully connected neural networks (FCNs) with convolutional neural networks (CNNs), as proposed by (Xie et al. 2024), are highly effective at extracting both basic and complex features from medical data. These hybrid models can handle large and diverse datasets efficiently by leveraging the strengths of CNNs for spatial data and FCNs for more abstract feature extraction, making them suitable for large-scale medical data,  demonstrating superior generalization capabilities, reaching an impressive accuracy of 94.8\%. \cite{xieDeepLearningApproach2024}. 

\subsubsection{Multiple Disease Prediction}

Selecting the right model is crucial for predicting multiple diseases, especially with diverse datasets and challenges in feature extraction, dataset management, performance, and generalizability. Among various approaches, those by (N. Nigar et al., 2023) and (R. Daid et al., 2021) are most effective for comorbidity prediction.

N. Nigar et al. (2023) advanced chronic disease diagnosis by combining five pretrained deep CNNs-VGG16, VGG19, ResNet, DenseNet, and Inception-v3. These models excel in feature extraction due to prior large-scale dataset training, discerning complex medical data patterns. Hyperparameter optimization and the XGBoost algorithm further enhance performance and accuracy. Data preprocessing, augmentation, and real-time datasets from Kaggle, integrated with the IoMT framework, ensure high performance and generalizability. However, success rates vary across diseases, indicating a need for additional fine-tuning for consistent results \cite{nigarIoMTMeetsMachine2023}.

Similarly, R. Daid et al. (2021) demonstrated exceptional performance with a deep learning-based CNN model trained on a dataset of 600,000 patients. This model effectively handles large data volumes and captures intricate disease patterns through hierarchical feature extraction, ensuring high predictive accuracy and generalizability across diseases. The model integrates various data types, including symptoms and environmental factors, for accurate predictions. Its performance, however, could be influenced by dataset representativeness and balance, requiring adaptability to diverse demographics and evolving medical conditions to maintain high performance  \cite{daidEffectiveMechanismEarly2021}.

\begin{figure}[htbp!]
    \centering
    \begin{subfigure}[b]{.5\textwidth}
        \centering
        \includegraphics[width=\textwidth]{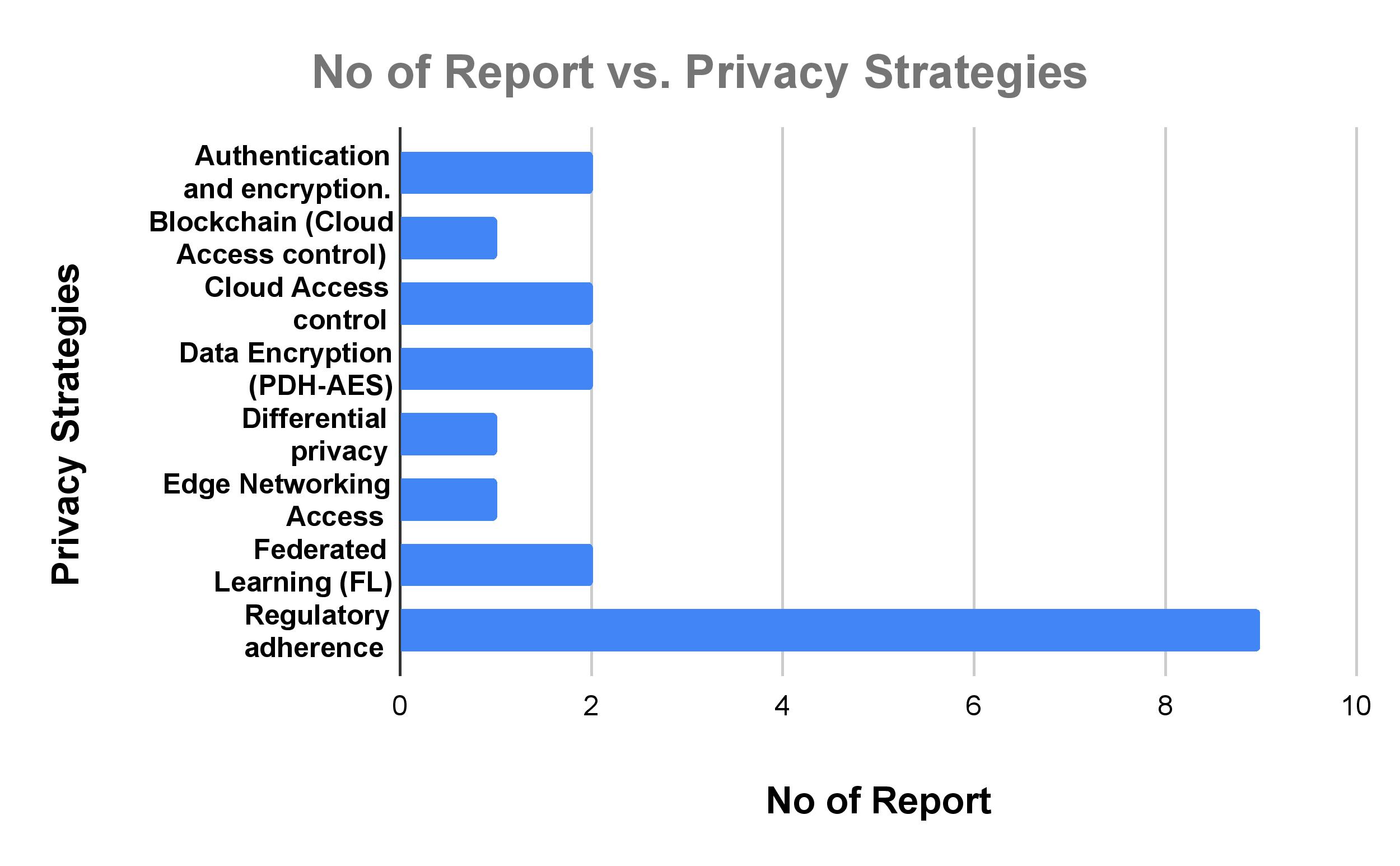}
        \caption{horizontal bar chart summarising IoMT-data privacy strategies reported in the review papers}
        \label{fig:privacy_strategies}
    \end{subfigure}
    \hfill
    \newline
    \begin{subfigure}[b]{.5\textwidth}
        \centering
        \includegraphics[width=\textwidth]{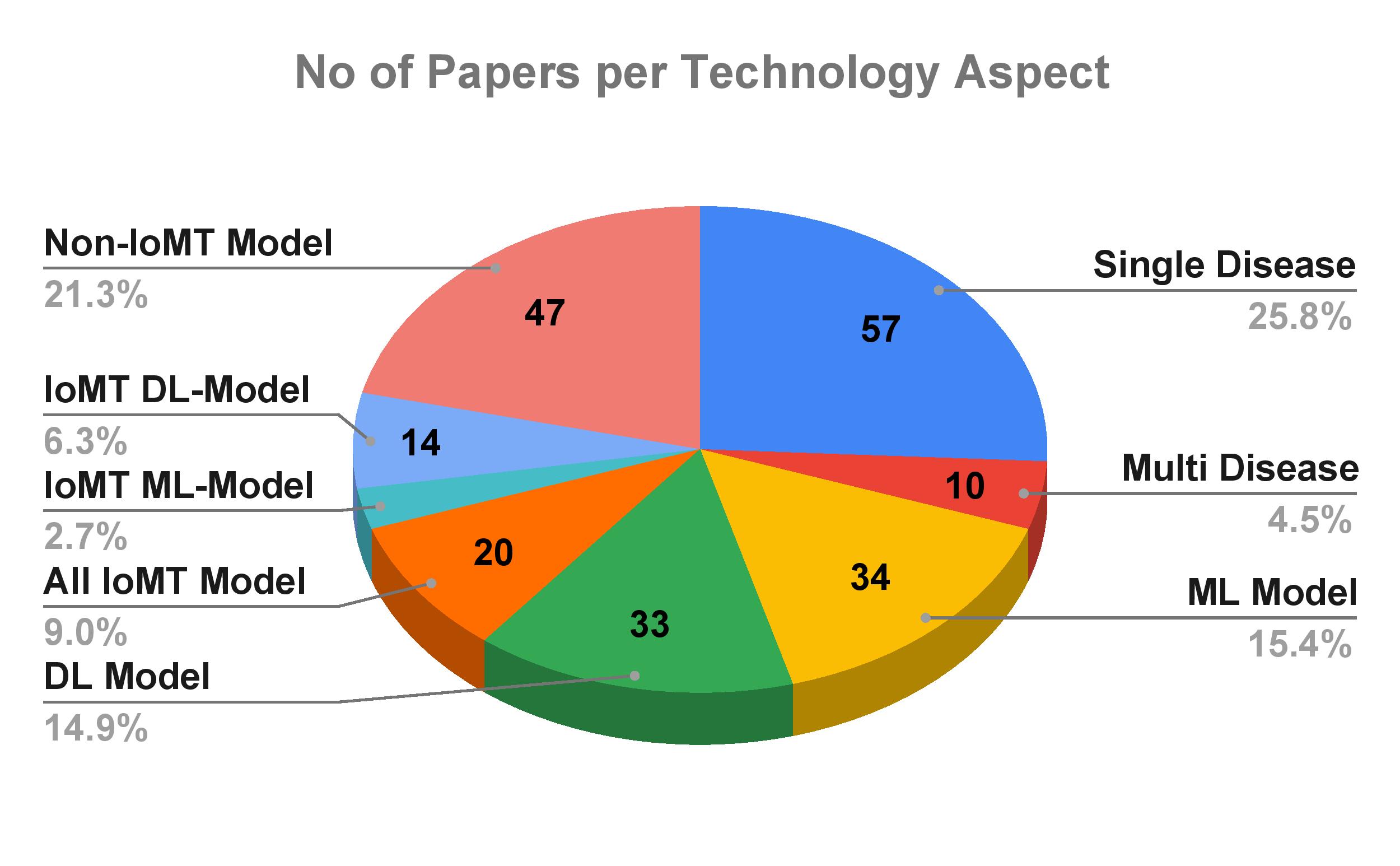}
        \caption{pie chart summarizing the number of reports for various aspect covered in the review}
        \label{fig:all_aspect_new}
    \end{subfigure}
    \caption{horizontal bar chart summarising IoMT-data privacy strategies reported in the review and pie chart summarizing the number of reports for various aspects covered in the review}
    \label{fig:all_aspects_diseases}
\end{figure}

\begin{figure}[!ht]
    \centering
    \includegraphics[width=\columnwidth]{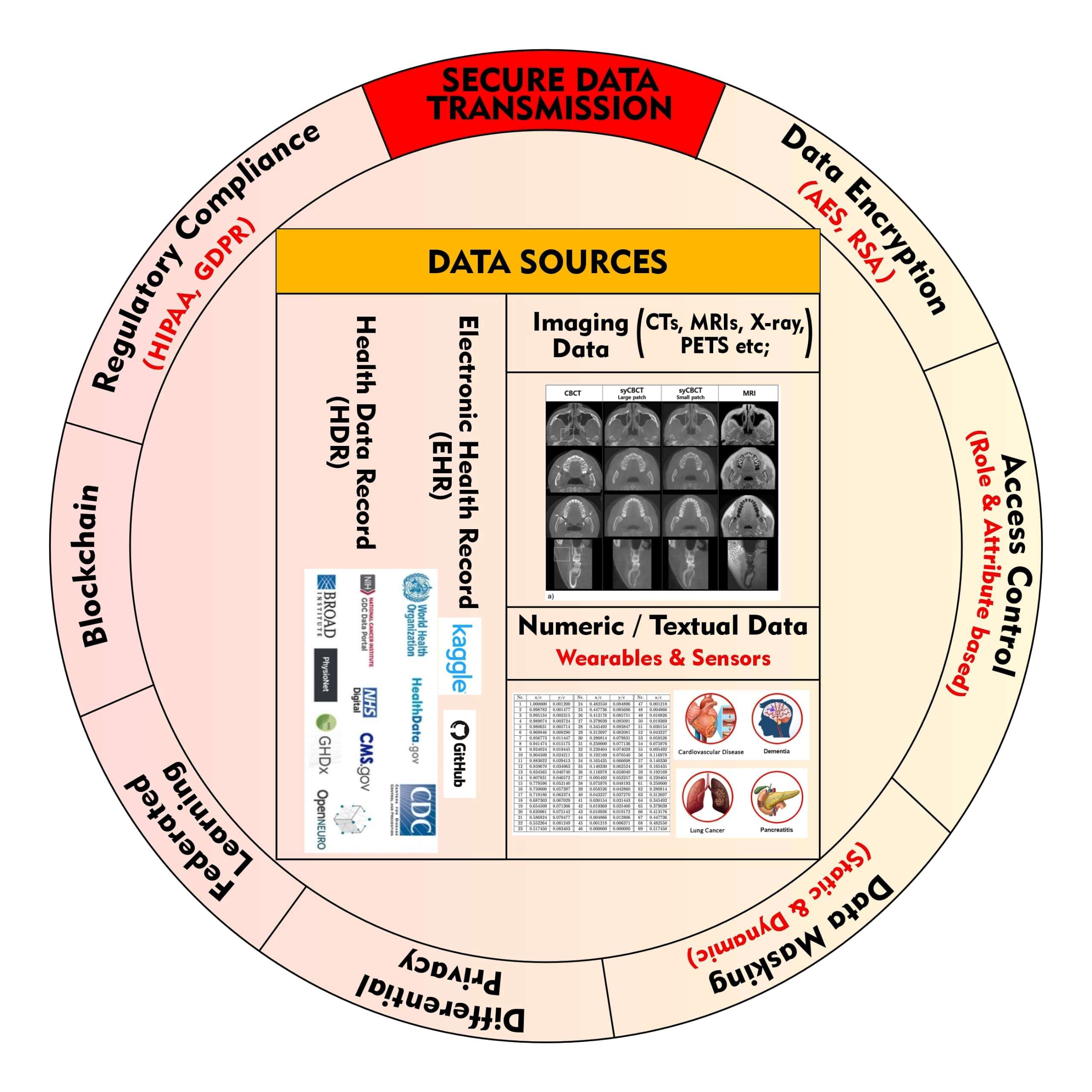}
    \caption{This illustration demonstrates the various data sources and the deployment of privacy protocols during data transmission for medical prognosis.}

    \label{fig:secure_data}
\end{figure}

Figure \ref{fig:all_aspects_diseases}. shows an horizontal bar chart summarising IoMT-data privacy strategies reported in the review papers and pie chart displaying the number of reports for each aspect of the review.  Additionally, Figure \ref{fig:secure_data} illustrates the various data sources and the deployment of privacy protocols during the transmission of data for medical prognosis.

\section{Discussion} \label{Sect4}

Artificial intelligence (AI) has significantly transformed medical diagnostics, offering unprecedented capabilities in the prediction and management of chronic and terminal diseases. By utilizing diverse and comprehensive datasets, AI models have improved in accuracy and generalizability, influencing clinical decision-making processes across various medical fields.

\paragraph{Dataset Challenges and Variability:}

The utilization of comprehensive datasets from reputable sources like the National Alzheimer’s Coordinating Center (NACC), Alzheimer’s Disease Neuroimaging Initiative (ADNI), and Open Access Series of Imaging Studies (OASIS-3) enables AI models to generalize across diverse populations and clinical conditions. However, the diversity and heterogeneity in these datasets present significant challenges:

1. Data Quality and Consistency: Differences in data collection protocols, patient demographics, and measurement standards can lead to variations in data quality. These inconsistencies can adversely affect model performance, resulting in unreliable outputs and compromised reproducibility across different healthcare settings. Incomplete or noisy data can further skew model training, potentially leading to biased predictions. Advanced data harmonization techniques, such as transfer learning and domain adaptation, are increasingly important. These methods help adapt models developed in one domain to perform effectively in another, thereby reducing performance degradation when applied to new data. Data imputation algorithms are also crucial for addressing issues related to missing data, ensuring that models are trained on comprehensive and representative datasets.

2. Overfitting Risks: Models trained in controlled environments with clean and curated data may exhibit high accuracy but often fail to replicate these results in real-world scenarios. This discrepancy, known as overfitting, occurs when models learn patterns specific to the training data, including noise, rather than generalizable features. To mitigate overfitting, robust validation frameworks are necessary. Techniques like k-fold cross-validation, adversarial validation, L1 and L2 regularization, dropout layers in neural networks, and early stopping during training can help prevent models from becoming overly complex and narrowly tailored to the training data.
Computational Complexity and Efficiency
While encryption and other security measures are essential for protecting sensitive patient data, they introduce additional computational overhead. For instance, encryption in models, such as Single-Layer Perceptrons for heart disease prediction, can significantly increase computational demands. This added complexity can hinder the scalability of AI solutions and limit their deployment in real-time clinical settings, where quick decision-making is essential.
To address this, homomorphic encryption can be utilized, allowing computations on encrypted data without decrypting it, thus preserving privacy while enabling robust data analytics. Additionally, lightweight cryptographic algorithms are beneficial in environments with limited processing power, such as SecureIoMT (Internet of Medical Things) devices, which collect and transmit sensitive health data.

\paragraph {Model Complexity and Deployment Challenges:}
Advanced models, like the two-stage architecture used by Manimurugan et al. for heart disease prediction, achieve high accuracy through sophisticated techniques. However, their complexity often requires substantial computational resources and expertise, which may not be available in all clinical settings. The complexity of these models can also make them difficult to interpret, posing challenges for integration into clinical workflows and decision-making processes.
To overcome these challenges, model simplification and knowledge distillation techniques can be employed, where simpler, more efficient models learn to replicate the performance of larger, more complex models. Additionally, edge computing offers a practical solution by processing data locally on SecureIoMT devices, reducing the need to transmit large volumes of sensitive data to centralized servers and enabling faster, real-time decision-making.

\paragraph{Data Security in Diverse Environments:}
Implementing robust security measures across varied healthcare systems is crucial for protecting patient data. The use of end-to-end encryption, zero trust architectures, and multi-factor authentication can significantly enhance data security. These measures ensure that data is protected from unauthorized access at all points, especially in environments utilizing SecureIoMT devices. Behavioral biometrics can further enhance security, ensuring that only authorized personnel can access critical data.
However, these security systems require significant infrastructure and maintenance, which can be challenging to implement uniformly across diverse healthcare settings. The diversity in healthcare environments, from advanced to resource-limited, adds complexity to the deployment of standardized security measures, potentially hindering the widespread adoption of AI technologies.

\paragraph{Mitigating Limitations for Broader Application:}
To ensure broader and more effective application of AI in healthcare, several strategies can be adopted:

1. Enhancing Data Interoperability: Improving the interoperability and integration of data from various sources, including SecureIoMT devices, is essential. These devices generate vast amounts of real-time data, necessitating sophisticated systems for efficient data management. Standards for data formatting and communication protocols are crucial for seamless data integration into AI models.

2. Improving Generalizability and Reliability: To combat overfitting and enhance model reliability, techniques such as cross-validation, robust regularization, and ensemble learning are important. These approaches ensure that models are accurate and applicable in diverse clinical environments, improving their robustness.

3. Standardizing Data Privacy Techniques: Consistent implementation of advanced data privacy methods, such as blockchain and differential privacy, is critical. These techniques protect patient data while ensuring compliance with regulations and maintaining trust. Standardization across healthcare systems facilitates data sharing and collaboration, enhancing the utility and reliability of AI models.

In conclusion, while AI holds promise in improving medical diagnostics through enhanced prediction capabilities, significant challenges such as dataset variability, overfitting and technical complexities must be addressed to ensure broader and more reliable application in clinical practice. Efforts to mitigate these limitations are crucial for advancing the effectiveness and scalability of AI-driven healthcare solutions.

\section{Gaps in the Literature} \label{Sect5}

 \paragraph{Data Quality and Generalizability:} 
AI models rely heavily on diverse datasets for training and validation. However, the variability in data quality, patient demographics, and data formats poses a significant challenge to the generalizability of these models. This variability is evident in datasets from different sources, including hospitals, clinical studies, and research initiatives like the Alzheimer's Disease Neuroimaging Initiative (ADNI) \cite{choiDeepLearningBasedStroke2021,el-sappaghMultilayerMultimodalDetection2021,wangPredictiveModelingProgression2018,jahanExplainableAIbasedAlzheimer2023}. The inconsistent data quality and diverse patient profiles can lead to models that perform well in specific settings but fail to generalize across different populations and clinical environments.

\paragraph{Complexity of Integrating IoMT Data:}
The rise of Internet of Medical Things (IoMT) devices, such as wearable health monitors and implantable sensors, introduces new layers of complexity. These devices continuously generate vast amounts of real-time patient data, which must be accurately integrated into AI models. The challenges lie in ensuring interoperability between different devices and systems, managing data standards, and handling the high volume of data effectively. The integration of IoMT data necessitates sophisticated data management strategies to ensure seamless incorporation into predictive models and clinical workflows.

\paragraph{Overfitting and Model Robustness:}
Overfitting remains a significant issue in the development of AI models for healthcare. Models often achieve high accuracy in controlled, curated environments but struggle to maintain performance when applied to new, unseen data in real-world clinical settings \cite{ogunpolaMachineLearningBasedPredictive2024,islamRiskFactorPrediction2020}. This gap highlights the need for robust model validation techniques and approaches to ensure that AI predictions are reliable and applicable in diverse clinical scenarios.

\paragraph{Addressing Multi-Morbidity Scenarios:}
While AI models have seen substantial development in predicting and diagnosing common diseases like cardiovascular disease and chronic kidney disease, there is a notable lack of focus on multi-morbidity scenarios, especially those involving rare and critical diseases such as dementia, stroke, and various cancers. These conditions often interact with other chronic illnesses, complicating diagnosis and treatment. The development of AI models that can effectively account for these interactions and provide accurate predictions in complex clinical scenarios remains underexplored and is critically needed.

\paragraph{Standardization and Optimization of Data Privacy Techniques:}
The protection of sensitive patient data is paramount, particularly in the context of AI-driven healthcare solutions that often rely on cloud-based infrastructures. However, there is an inconsistent implementation of advanced data privacy techniques such as federated learning, blockchain, encryption, and differential privacy across healthcare systems. This inconsistency leads to variations in data protection standards and computational efficiency. The integration of IoMT devices further complicates the scenario, as these devices continuously collect sensitive patient data. Ensuring robust security measures, including secure transmission protocols and endpoint security, is essential to prevent unauthorized access and maintain patient confidentiality. By addressing these challenges comprehensively, healthcare systems can enhance trust in AI-driven technologies while safeguarding patient data integrity and regulatory compliance.

\section{Future Research Directions} \label{Sect6}
As the field of predictive modeling in healthcare continues to evolve, several key areas present promising opportunities for advancement. Addressing these areas can significantly enhance the efficacy and applicability of predictive models in clinical settings. Future research should focus on the following directions:

1. Exploration of Multi-Disease Models: There is a significant opportunity to develop predictive models that can address multiple diseases concurrently, particularly in environments characterized by prevalent comorbidities. This includes focusing on multi-morbidity scenarios, such as the interaction between rare and critical diseases like dementia, stroke, and various cancers, and their intersection with chronic illnesses.

2. Enhancement of Data Interoperability: Effective predictive models rely on the quality and integration of data from diverse sources. To address this, advanced data preprocessing techniques are essential. This includes implementing robust protocols for data cleaning, transformation, and harmonization to standardize data formats and quality, especially when integrating data from the Internet of Medical Things (IoMT) devices.

3. Focus on Model Explainability: As predictive models grow in complexity, maintaining their interpretability for practitioners becomes crucial. Research should aim at developing methods to make these sophisticated models more understandable, thereby bridging the gap between technical efficacy and clinical usability. This includes addressing overfitting and generalizability challenges by employing techniques like transfer learning and ensemble methods, which can improve model performance across various data sources, including those generated by IoMT devices.

4. Integration of Advanced Data Privacy Techniques: Standardizing the implementation of advanced techniques such as federated learning, blockchain, and differential privacy is imperative. This can be achieved through the creation of open-source tools and frameworks to simplify the integration of these methods into IoMT systems, aiding in the management of large datasets from diverse devices. The effort should focus on scaling these techniques to facilitate real-time, collaborative analysis across decentralized networks while ensuring data integrity and transparency.

\section{Conclusion} \label{Sect7}

The fusion of machine learning (ML) and deep learning (DL) with the Internet of Medical Things (IoMT) has significantly advanced the field of medical diagnostics. ML algorithms like XGBoost and Random Forest are pivotal in achieving high accuracy levels, noted at over 98\% for diseases such as heart disease, chronic kidney disease (CKD), Alzheimer’s, and lung cancer. These algorithms excel in handling structured data, performing well in environments where large volumes of historical data aid in predictive analysis.

DL techniques, including Convolutional Neural Networks (CNNs) and Long Short-Term Memory Recurrent Neural Networks (LSTM RNNs), have shown exceptional promise in interpreting complex patterns in data, particularly useful in forecasting conditions like stroke and dementia. Their high accuracy, exceeding 99\% in some studies, is largely due to their proficiency in processing sequential and image data, which is often gathered in real-time from various medical sensors integrated within IoMT frameworks.

These breakthroughs leverage cloud-based analytics and real-time data from medical sensors, enhancing the ability to predict and diagnose diseases such as kidney and lung cancer. This integration not only improves the speed and accuracy of diagnostics but also helps in continuous patient monitoring, crucial for chronic and terminal disease management.

Despite these technological advancements, the integration of AI and IoMT is not without challenges. Variability in data quality and sources, often compounded by the complexities inherent in IoMT devices, presents significant hurdles. These include: (i)
Data Management and Integration: The diversity of data types and sources requires robust data management and integration systems. Effective handling of this data is crucial to ensure that the predictive models are both accurate and scalable across different healthcare settings; (ii) Overfitting and Generalization: Overfitting remains a significant challenge, where models trained on specific datasets perform well in controlled tests but fail to generalize to broader, real-world applications. This issue is particularly pronounced in the context of multi-morbidity scenarios and rare diseases, where data can be sparse and highly specific; (iii) Navigating Multi-Morbidity Scenarios: Diseases often do not occur in isolation. The ability to predict and diagnose multiple conditions simultaneously, or to understand how one disease may influence the progression of another, is still an area needing extensive research and innovative modeling approaches.

Ensuring the privacy and security of patient data is paramount. With the increasing use of IoMT devices that continuously collect patient data, establishing standardized and optimized privacy measures is critical to maintaining patient trust and ensuring compliance with stringent healthcare regulations. The challenges include: (i)
Standardization Across Platforms: Uniform data privacy standards that can be implemented across various platforms and devices are essential. This standardization helps in safeguarding data integrity and ensuring seamless interoperability between different systems; (ii) Optimized Data Privacy Measures: Employing advanced techniques such as federated learning, blockchain technology, and differential privacy can help in enhancing the security and privacy of patient data, minimizing the risk of breaches while maintaining the utility of the data for medical analysis.

The graphical abstract depicted on the first page of our review paper highlights essential elements, including dataset availability, privacy protocols, and the use of machine learning (ML) and deep learning (DL) models such as XGBoost, CNNs, and LSTM networks for predicting conditions like heart disease, CKD, Alzheimer's, and lung cancer. It explores the integration of diverse datasets from platforms like Kaggle, UCI, private institutions, and IoMT sources. The figure also addresses issues related to data interoperability, model generalization, and the necessity for advanced methods like transfer learning and ensemble techniques, while emphasizing the need for strong privacy and security measures for IoMT systems. Additionally, it outlines future research directions, provides an analysis of each section, identifies existing gaps, and proposes future directions.

\section* {CRediT Authorship Contribution Statement}
Akeem Temitope Otapo (Doctoral Student) Conceptualization, Methodology, Investigation, Formal Analysis, Writing - Original Draft, and Data Curation. Alice Othmani (Corresponding Author, Supervisor) was responsible for Supervision, Writing - Review \& Editing, and Administration. Ghazaleh Khodabandelou (Co-Supervisor) Contributed to conceptualization, Validation, and Writing - Review \& Editing. Zuheng Ming (Co-Supervisor) contributed to Conceptualization, Methodology, Resources, and Visualization.

\section* {Declaration of Competing Interest}
The authors declare that they have no known competing financial interests or personal relationships that could have appeared to influence the work reported in this paper.

\section* {Acknowledgments}
The authors would like to acknowledge Kazeem Adetunji Sodiq and Fati Oiza Salami for their invaluable assistance with technical support and manuscript preparation. Special thanks to Yahaya Idris Abubakar for the insightful discussions that significantly enhanced the quality of the manuscript.

\section*{Preprint License} This article is a preprint and has not been peer-reviewed. It is made available under a CC-BY-NC-ND license, which permits non-commercial use, distribution, and reproduction in any medium, provided the original work is properly cited, and no modifications or adaptations are made. For full license details, visit https://creativecommons.org/licenses/by-nc-nd/4.0/.

\section* {Funding}
This work is supported by the Tertiary Education Trust Fund (TETFund), Nigeria, through the provision of a study grant that made this research possible.

\section* {Declaration of generative AI and AI-assisted technologies in the writing process}
During the preparation of this work, the authors used [ChatGPT and Quillbot] to [improve the grammar and readability of the text]. After utilizing these tools/services, the authors reviewed and edited the content as needed and take full responsibility for the final version of the published article.

\onecolumn
\begin{longtable}{p{3.5cm} p{1.5cm} p{1.5cm} p{1.2cm} p{1.2cm} p{1.5cm} p{1.cm} p{2.5cm}}
\caption{This table provides an overview of the available datasets used in research on various chronic and terminal diseases, including Alzheimer's, arrhythmia, liver disease, pancreatic disease, breast cancer, and more. For each dataset, the table lists the reference, disease area, dataset name, number of participants, number of samples, data type, source, and use in research. Legend: - for Not Provided.}
\label{tab:used_datasets} \\
\toprule
\textbf{Ref.} & \textbf{Disease} & \textbf{Dataset} & \textbf{Participants} & \textbf{Samples} & \textbf{Modality} & \textbf{Source} & \textbf{Use in Research} \\
\midrule
\endfirsthead
\caption[]{(continued)} \\
\toprule
\textbf{Ref.} & \textbf{Disease} & \textbf{Dataset} & \textbf{Participants} & \textbf{Samples} & \textbf{Modality} & \textbf{Source} & \textbf{Use in Research} \\
\midrule
\endhead
\midrule
\endfoot
\bottomrule
\endlastfoot

Pamela LaMontagne et al. (2019) \cite{lamontagneOASIS3LongitudinalNeuroimaging2019} & Alzheimer & OASIS-3 & 1098 & 3500 & Images & \href{https://doi.org/10.1101/2019.12.13.19014902}{link} & Alzheimer's Diagnostic \cite{jahanExplainableAIbasedAlzheimer2023}\\
\hline
Suhuai et al. (2017) \cite{luoAutomaticAlzheimerDisease2017} & Alzheimer & ADNI & 81 & 81 & Images & \href{https://doi.org/10.4236/jamp.2017.59159}{link} & Alzheimer's Diagnostic \cite{choudhuryCoupledGANArchitectureFuse2024}\\
\hline
StPetersburg INCART (2008) \cite{tihonenkoStPetersburgInstitute2007} & Arrhythmia & ECG & 32 & 75 & Tabular & \href{https://doi.org/10.13026/C2V88N} {link} & Arrhythmia classification \cite{sharmaComprehensiveReviewArrhythmia2023}\\
\hline
MIT-BIH Arrhythmia \cite{moodyMITBIHArrhythmiaDatabase1992} & Arrhythmia & ECG & 47 & 48 & Tabular & \href{https://doi.org/10.13026/C2F305} {link} & Arrhythmia detection \cite{walinjkarSignalProcessingEarly2019}\\
\hline
Karoly et al. (2021) \cite{karolyMultidayCyclesHeart2021} & Epilepsy & Heart rate & 46 & 46 & Tabular & \href{https://doi.org/10.1016/j.ebiom.2021.103619} {link} & Cardiac Monitoring \cite{nurseMultidayCyclesHeart2022}\\
\hline
Kenneth Clark et al. (2013) \cite{clarkCancerImagingArchive2013} & Cancer & Cancer & 37568 & 2074 & Images & \href{https://doi:10.1007/s10278-013-9622-7} {link} & Breast cancer diagnosis \cite{carrieroDeepLearningBreast2024}\\
\hline
Soundarapandian et al. (2015) \cite{l.rubiniChronicKidneyDisease2015} & CKD & CKD & - & 400 & Tabular & \href{https://doi.org/10.24432/C5G020} {link} & Kidney disease diagnosis \cite{alsekaitComprehensiveChronicKidney2023}\\
\hline
Yazeed Zoabi et al. (2024) \cite{nshomronNshomronCovidpred2024} & Covid & Multivariate & 99232 & 99232 & Tabular & \href{https://doi.org/10.1038/s41746-020-00372-6} {link} & Covid-19 Research \cite{mardianReviewCurrentCOVID192021}\\
\hline
Johnson et al. (2016) \cite{johnsonMIMICIIIFreelyAccessible2016} & Critical care & MIMIC III & 46520 & 61532 & Tabular & \href{https://doi.org/10.13026/C2XW26} {link} & Intensive care research \cite{daiAnalysisAdultDisease2020}\\
\hline
Bendi Ramana et al. (2012) \cite{bendiramanaILPDIndianLiver2012} & ILPD & Multivariate & 10 & 583 & Tabular & \href{https://doi.org/10.24432/C5D02C} {link} & Liver Disease Research \cite{abdalradaPredictiveModelLiver2019}\\
\hline
Andras Janos (1989) \cite{andrasjanosiHeartDisease1989} & Heart & Multivariate & - & 303 & Tabular & \href{https://doi.org/10.24432/C52P4X} {link} & Heart Disease Study \cite{hussainNovelDeepLearning2021}\\
\hline
Mysar Ahmad Bhat (2021) \cite{mysarLungCancer2021} & Lung Cancer & Multivariate & 284 & 16 & Tabular & \href{https://www.kaggle.com/datasets/mysarahmadbhat/lung-cancer} {link} & Lung cancer research \cite{p.r.ComparativeStudyLung2019}\\
\hline
Golovenkin et al. (2020) \cite{s.e.golovenkinMyocardialInfarctionComplications2020} & Myocardial infarction & CAD & - & 1700 & Tabular & \href{https://doi.org/10.24432/C53P5M} {link} & Cardiac Research \cite{doudesisMachineLearningDiagnosis2023}\\
\hline
Holger R. Roth et al. (2015) \cite{rothNIHPancreasCTDataset2017} & Pancreas CT & CT & 82 & 82 & Images & \href{https://doi.org/10.7937/K9/TCIA.2016.tNB1kqBU} {link} & Pancreatic cancer diagnosis \cite{liAccuratePancreasSegmentation2021}\\
\hline
Vikas Ukani (2009) \cite{vikasukaniParkinsonDiseaseData2009} & Parkinson & Multivariate & 31 & 195 & Tabular & \href{https://www.kaggle.com/datasets/vikasukani/parkinsons-disease-data-set} {link} & Parkinson's Disease \cite{littleSuitabilityDysphoniaMeasurements2009}\\
\hline

\end{longtable}

\begin{longtable} {p{1.2cm}p{1.2cm}p{3cm}p{3cm}p{3cm}p{1.5cm}p{1.5cm}}
\captionsetup{width=\textwidth}
\caption{Summarizes all the thirty-four studies that apply various ML techniques such as RF, DT, LR, XGBoost, SVM, KNN, NB, LGBM, GBDT, ANN, etc., to predict and analyze chronic diseases. Each study is categorized by disease type and ML model used, with reported performance metrics such as Accuracy, Sensitivity, Specificity, ROC AUC, and F1-score. Dataset descriptions encompass diverse sources from medical imaging to clinical records. Notably, integration with Internet of Medical Things (IoMT) data and stringent data privacy measures are highlighted across studies to protect patient information. Legend: - Not Provided} \label{tab:ML_publications} \\\hline

\textbf{Ref.} & \textbf{Disease Type} & \textbf{ML Model} & \textbf{Model Performance} & \textbf{Dataset Description} & \textbf{IoMT} & \textbf{Privacy} \\\hline
\endfirsthead
\multicolumn{7}{c}%
{\tablename\ \thetable\ -- \textit{Continued from previous page}} \\\hline
\textbf{Ref.} & \textbf{Disease Type} & \textbf{ML Model} & \textbf{Model Performance} & \textbf{Dataset Description} & \textbf{IoMT} & \textbf{Privacy} \\ \hline
\endhead
\hline \multicolumn{7}{r}{\textit{Continued on next page}} \\\endfoot

\hline
\endlastfoot
C. Zhang et al. (2018) \cite{zhangPPDPEfficientPrivacypreserving2018} & Heart Disease & Privacy-Preserving data in the cloud and SLP classifier & 16.83\% for Breast Cancer, 16.13\% for Heart Disease & UCI Heart Disease (297 samples, 13 attributes), Breast Cancer (683 samples, 9 attributes), and Synthetic Dataset & Cloud-based e-Healthcare & Data Encryption \\\hline
K. S. Adewole et al. (2021) \cite{adewoleChapterFiveCloudbased2021} & CVD (Heart) & IoMT Cloud-based (CPHM) SVM with linear, polynomial, and radial basis kernels, NB and KNN algorithms & SVM (73.75\%), NB (91.07\%), KNN (65.41\%) & Kaggle dataset (70,000 patients, 12 attributes) & Cloud-based e-Healthcare & Digital watermarking with cloud security \\ \hline 

X. Yuan et al. (2022) \cite{yuanStableAIBasedBinary2022} & Heart Disease & Stable AI-Based Binary integrates the Bagging-Fuzzy-GBDT algorithm & 80-95\% accuracy & UCI dataset (836 samples, 14 attributes) & IoMT sensors & Differential privacy in tree-based data mining \\ \hline

A. Ed-daoudy et al. (2024) \cite{ed-daoudyScalableRealtimeSystem2023} & Heart Disease & Four-tier Architecture with Spark MLlib & RF (92.05\%), LG/SVM (88.64\%), MLP (87.50\%), DT (79.55\%) & UCI dataset (14 attributes) & Wearable sensors & -  \\ \hline

D. Asif et al. (2023) \cite{asifEnhancingHeartDisease2023} & Heart Disease & Ensemble Learning integrating extra tree classifier, RF, XGBoost, and CatBoost algorithms focusing on HP optimization & Extra Tree 97.23\% to 98.15\% accuracy & Merged datasets from Kaggle (297, 1025, 303 samples) & - & - \\ \hline

A. Ogunpola et al. (2024) \cite{ogunpolaMachineLearningBasedPredictive2024} & CVD (Heart) & Seven Machine Learning and Deep Learning Classifiers & XGBoost (98.50\% accuracy) & Mendeley (1,000 samples, 13 attributes), Kaggle (303 samples, 14 attributes) & - & - \\ \hline

E. A. Radya and A. S. Anwar (2019) \cite{radyPredictionKidneyDisease2019} & Kidney Disease Stage & Mining algorithms PNN, MLP, SVM, and RBF & PNN algorithm: Excellent Accuracy Precision and F-Measure across all the stages & UCI 361 CKD Indian patients and included 25 variables & - & -  \\ \hline

Md. A. Islam et al. (2020) \cite{islamRiskFactorPrediction2020} & Risk Factor (CKD) & ML algorithms like DT, NB, RF, and with LR to calculate associated risk factors & Accuracies NB (93.91\%), RF (98.89\%), and LR (94.77\%). Hemoglobin as most significant risk factor & 1032 questionnaire-based patient records medical college in Bangladesh & - & - \\ \hline

Z. Segal et al. (2020) \cite{segalMachineLearningAlgorithm2020} & End-Stage Renal Disease & Employs machine-learning techniques, particularly the XGBoost algorithm & Strong performance with a C-statistic of 0.93, sensitivity of 0.715, specificity of 0.958, PPV of 0.517, and NPV of 0.981 & Medical insurance data claims from 550,000 CKD patients over 18 years & - & -  \\ \hline

P. Ventrella et al. (2021) \cite{ventrellaSupervisedMachineLearning2021} & Chronic Kidney Disease Advancement & Employs machine-learning techniques. ExtraTrees classifier excelled & Accuracies of 0.94 for 2 classes, 0.91 for 3 classes, and 0.87 for 4 classes & Data from Vimercate Hospital's Electronic Medical Records (EMR) & - & -  \\ \hline

K. H. Lee et al. (2022) \cite{leeArtificialIntelligenceRisk2022} & End-Stage Renal Disease & Various machine learning algorithms, RF, Extra Trees, XGBoost, LGBM, and GBDT & GBDT achieving the highest AUC of 0.879 & 112,628 CKD (Chronic Kidney Disease) patients using ICD-9 and ICD-10 diagnostic codes & - & - \\ \hline

Murugesan G et al. (2022) \cite{g.FuzzyLogicBasedSystems2022} & Chronic Kidney Disease & Fuzzy logic principles with ANN capabilities & Achieved an accuracy of 93.75\% with high sensitivity, specificity, and precision rates & Dataset of various patients with chronic kidney disease was collected from expertise or specialist doctors of nephrology & - & - \\ \hline

Z. B. Miled et al. (2020) \cite{benmiledPredictingDementiaRoutine2020} & Dementia Prediction & RF model was employed using routine care electronic medical records (EMR) data & Generalize with 77.43\% accuracy, 76.01\% sensitivity, and 74.16\% specificity & 19 dementia-related disorders identified using ICD-10 or ICD-9 codes. Total of 104 features in the Rx data and 23 features in the Dx data & - & - \\ \hline

Yingjie Li et al. (2020) \cite{liEarlyPredictionAlzheimer2020} & Alzheimer & PACE for extraction of functional principal component  & Sens. 80\%, Spec. 70\%, Acc. 75\%, AUC 80\%. EC best predictor (AUCs above 0.83 for 1-year and 2-year) & ADNIMERGE file MRI data with clinical info for MCI & - \\ \hline

S. El-Sappagh et al. (2021) \cite{el-sappaghMultilayerMultimodalDetection2021} & Alzheimer’s Disease & Two-layered model with RF classifiers and explainers predicts cognitive states and Alzheimer's progression using DT and Fuzzy Rule-Based Systems for interpretability & Cross-validation accuracy of 93.95\% and an F1-score of 93.94\% in the first layer, and a cross-validation accuracy of 87.08\% and an F1-Score of 87.09\% in the second layer & ADNI includes data from 1048 subjects & - & -  \\ \hline

Chatterjee et al. (2022) \cite{chatterjeeVotingEnsembleApproach2022} & Alzheimer & SVM, KNN, Logistic Regression, Naive Bayes & Accuracy: 96.43\% & OASIS project (longitudinal and cross-sectional MRI, 150 patients) & - \\ \hline

Khan et al. (2022) \cite{khanEnsembleModelDiagnostic2022} & Alzheimer & KNN, NB, GB, XGB, SVM, ensemble of XGB, DT, SVM & Efficiency: 95.75\% & 2127 MRI (T1 and T2) from ADNI, 612 AD, 538 MCI, 975 CN & - \\ \hline

S. Jahan et al. (2023) \cite{jahanExplainableAIbasedAlzheimer2023} & Alzheimer’s Disease & For predicting five class classifications, 9 most popular Machine Learning models & Accuracy: 94.51\% to 98.94\%, Precision: 82.35\% to 96.32\%, Recall: 94.41\% to 98.79\%, F1-score: 81.92\% to 98.75\%, AUC: 53.4\% to 99.97\% & (OASIS)-3 1098 unique participants for clinical data, 1053 for MRI segmentation data, and 810 for psychological assessment data & Wearable sensor bands worn by Alzheimer's disease (AD) patients & A mist layer at the network's edge, handling tasks like filtering and aggregation \\ \hline

W. Muhammad et al. (2019) \cite{muhammadPancreaticCancerPrediction2019} & Pancreatic Cancer & ANN model for predicting pancreatic cancer risk & Sensitivity and specificity of 80.7\%, PPV of 0.089\%, NPV of 99.995\%, and AUC of 0.85 & Combined NHIS and PLCO datasets with 800,114 participants and 898 cases & - & -  \\ \hline

Y. Zhou et al. (2022) \cite{zhouPredictionSeverityAcute2022} & Severity of Acute Pancreatitis & Uses 5 algorithms (LR, SVM, DT, RF, XGBoost) & XGBoost outperformed with an AUC of 0.906, an accuracy of 0.902, a sensitivity of 0.700, a specificity of 0.961, and an F1 score of 0.764 & 441 patients were included in this study & - & - \\ \hline

B. Kui et al. (2022) \cite{kuiEASYAPPArtificial2022} & Prediction of Severity in Acute Pancreatitis & Used machine learning algorithms, DT, RF, LR, SVM, CatBoost, and XGBoost, for binary classification of acute pancreatitis severity & XGBoost classifier, exhibiting an average AUC score of 0.81, accuracy of 89.1\% & International cohort of 1184 patients for model development and a validation cohort of 3543 patients & - & - \\ \hline

M. Moll et al. (2020) \cite{mollMachineLearningPrediction2020} & COPD & Top features in a Cox regression to create a machine learning mortality prediction (MLMP) in COPD using RF & The MLMP-COPD model exhibited superior predictive performance achieving a C index of at least 0.7 & Using 2,632 participants from the COPDGene Study and validated on 1,268 participants from the ECLIPSE Study & - & - \\ \hline

S. Rajesh et al. (2020) \cite{rajeshHepatocellularCarcinomaHCC2020} & (HCC) Liver Cancer & Five ML algorithms KNN, NB, DT, RF, and SVM employing k-Fold cross-validation & Highest accuracy with RF 79.46\% & Utilized two datasets, HCC STo from the UCI and HCC Dataset 3 from Kaggle & - & - \\ \hline

Barus et al. (2022) \cite{barusLiverDiseasePrediction2022} & Liver & SVM, LR, PCA, SMOTE & LR: 70\%, SVM: 88\% before PCA/SMOTE; LR: 64\%, SVM: 87\% after PCA/SMOTE & ILPD: 583 records, 11 features & - & -\\ \hline

Md et al. (2023) \cite{mdEnhancedPreprocessingApproach2023} & Liver & GB, XGB, Bagging, RF, ET, Stacking & Accuracy: 91.82\%, 86.06\% & ILPD: 583 records, 11 features & - & -\\ \hline

Ruhul Amin et al. (2023) \cite{aminPredictionChronicLiver2023} & Liver Disease & Combines FA and LDA for feature selection, utilizes ML techniques like RF, KNN, LR, MLP, SVM, and BRT & Accuracy: 88.10\%, Precision: 85.33\%, Recall: 92.30\%, F1 Score: 88.68\%, AUC: 88.20\% & The Indian Liver Patient Dataset (ILPD) - UCI consists of 583 observations, 10 features & - & - \\ \hline

Kumar et al. (2023) \cite{kumarLiverDiseasePrediction2023} & Liver & DT, KNN, MLP, AB, RF, GB, XGB, LR, NB, ET, LGBM, SVM & DT Accuracy: 86.67\%, Precision: 0.87, Recall: 0.87, F1: 0.86 & ILPD & - &-\\ \hline

Warda M. Shaban (2023) \cite{wardam.shabanEarlyDiagnosisLiver2023} & Liver Disease & Liver Patients Detection Strategy (LPDS) using ML classifiers like SVM, KNN, NB, DT, and RF & Accuracy: 89.5\%, Sensitivity: 91.2\%, Specificity: 87.3\%, F1 Score: 0.89, AUC: 0.92. KNN classifier achieved 99.1\% on the test dataset & Kaggle Liver dataset total number 441 & - & - \\ \hline

S. Abbas et al. (2023) \cite{abbasFusedWeightedFederated2023} & Lung Cancer Disease & Fused Weighted Federated Deep Extreme Machine Learning Based on Intelligent Lung Cancer Disease Prediction & Achieved 96.3\% which is better than the state-of-the-art method & The lung cancer 309 cases initially sourced from sensors, supplemented by 231 records for dataset equalization & Data transmitted to a raw database using IoMT technologies & Federated Learning (FL) as a data privacy measure \\ \hline

N. K. Sbehara et al. (2015) \cite{beheraBirdMatingOptimization2015} & Thyroid, Hepatitis, Heart Diseases & Multilayer Perceptron (MLP) with Bird Mating Optimization (BMO) and Firefly Algorithm (FFA) & MLP-BMO outperformed avg: 2.9802, std: 0.3534 & Biomedical datasets like thyroid, hepatitis, heart diseases, liver diseases, and Indian Pima diabetes disease (UCI) & - & - \\ \hline

Akkem Yaganteeswarudu (2020) \cite{yaganteeswaruduMultiDiseasePrediction2020} & Diabetes, Heart Disease, Cancer, Diabetic Retinopathy & ML, TF, and Flask API with parameter analysis for optimal disease detection, employing LR, NB, SVM, DT, RF achieving high accuracy rate & Accuracies: diabetes 92\% with LR, 95\% heart with RF, and 96\% cancer with SVM & PIDD for diabetes, 150 GB UCI ML data for DR, HD data from Cleveland, Hungarian, Switzerland, BCWD for cancer, hospital live data for parameters & - & - \\ \hline

P. Singh et al. (2021) \cite{singhMultidiseaseBigData2021} & Heart Disease and Multi-Disease & Beetle Swarm Optimization and Adaptive Neuro-Fuzzy Inference System (BSO-ANFIS) & 99.1\% accuracy, 99.37\% precision; Multi-disease classification: 96.08\% accuracy, 98.63\% precision & Heart disease dataset from Kaggle. Multi-disease dataset from USA healthcare and services & - & - \\ \hline

J. Rash et al. (2022) \cite{rashidAugmentedArtificialIntelligence2022} & Breast Cancer, Diabetes, Heart Attack, Hepatitis, Kidney Disease & ANN with PSO & Highest accuracy: 99.67\% for chronic disease prediction, outperforming LR, SVM, KNN, NB, DT. Feature optimization yielded 99.65\% accuracy & Datasets from Kaggle, Dataworld, Github, UCI & - & - \\ \hline

\end{longtable}

\begin{longtable}
{p{1.2cm}p{1.2cm}p{3cm}p{3cm}p{3cm}p{1.5cm}p{1.5cm}}
\captionsetup{width=\textwidth}
\caption{Summarizes thirty-three studies that apply various deep learning (DL) techniques, including VGG19 CNN, NN, RNN, DBN, LSTM, GoogleNet, VGG16 CNN, ResNet, and Inception, to predict and analyze chronic diseases. Each study is categorized by disease type and ML model used, with reported performance metrics such as Accuracy, Sensitivity, Specificity, ROC AUC, and F1-score. Dataset descriptions encompass diverse sources from medical imaging to clinical records. Notably, integration with Internet of Medical Things (IoMT) data and stringent data privacy measures are highlighted across studies to protect patient information.
LEGEND: - Not Provided} \label{tab:DL_publications} \\
\hline
\textbf{Ref.} & \textbf{Disease Type} & \textbf{ML Model} & \textbf{Model Performance} & \textbf{Dataset Description} & \textbf{IoMT} & \textbf{Privacy}  \\ \hline
\endfirsthead
\multicolumn{7}{c}%
{\tablename\ \thetable\ -- \textit{Continued from previous page}} \\
\hline
\textbf{Ref.} & \textbf{Disease Type} & \textbf{ML Model} & \textbf{Model Performance} & \textbf{Dataset Description} & \textbf{IoMT} & \textbf{Privacy}  \\ \hline
\endhead

\hline \multicolumn{7}{r}{\textit{Continued on next page}} \\
\endfoot
\hline
\endlastfoot
Y. PAN et al. (2020) \cite{panEnhancedDeepLearning2020} & Heart Disease & Enhanced Deep Convolutional Neural Network (EDCNN) & Accuracy result of 97.51\% & UCI clinical test dataset & collect parameters like pulse, ECG, and blood pressure & - \\\hline
S.S. Sarmah (2020) \cite{sarmahEfficientIoTBasedPatient2020} & Heart Disease & Deep Learning Modified Neural Network (DLMNN) & DLMNN  outperforms ANN with 92\% accuracy & Hungarian Heart Disease (HD) dataset, 294 records & Body sensor  & cipher-based hash code for user Verification \\\hline
S. Hussain et al. (2021) \cite{hussainNovelDeepLearning2021} & Heart Disease & 1D Convolutional Neural Network (CNN) & Achieved over 97\% training accuracy and 96\% test accuracy & Cleveland heart disease dataset containing 303 samples & - & -  \\
\hline
A. Kumar et al. (2022) \cite{kumarECGBasedEarly2022} & Early Heart Attack & Convolutional Neural Network (CNN) & achieved 98\% accuracy, precision of 97\%, and average F-score of 98\% & UCI dataset contains 383 data points and 14 features & - & - \\
\hline
S. M. Nagarajan et al. (2022) \cite{nagarajanInnovativeFeatureSelection2022} & Heart Disease & GCSA model integrated with DCNN &  obtained 94\% classification accuracy original features and 95.34\% for extracted features & Ten different medical datasets  & - & - \\
\hline
S. Manimurugan et al. (2022) \cite{manimuruganTwoStageClassificationModel2022} & Heart Disease & HLDA-MALO technique and Faster R-CNN with SE-ResNet-101 & HLDA 96.85\% acc. normal and 98.31\% abnormal sensor data , Faster R-CNN with SE-ResNet-101 max. acc. of 99.15\% for ECG image & Cleveland dataset  and ECG image from UCI  & collect sensor data & by storing data in a cloud database \\
\hline
A. A. Nancy et al. (2022) \cite{nancyIoTCloudBasedSmartHealthcare2022} & Heart Disease & Bi-LSTM with fuzzy system &  model achieves accuracy values ranging from 98.03\% to 98.90\% & Cleveland and Hungarian datasets UCI &  remote patient monitoring & Data privacy through encryption and access control \\\hline
L. Kumar et al. (2023) \cite{kumarDeepLearningBased2023} & Heart Disease & CNN-based model & Achieved 98.5\% accuracy  & Dataset includes 2,976 individuals with heart failure & - &- \\\hline
S. Deepa et al. (2023) \cite{sExperimentalEvaluationArtificial2023} & Cardiovascular Disease & Efficient Learning Health Evaluator (ELHE) & Achieved over 97\% accuracy & UCI laboratory datasets & - & - \\
\hline
M. Munsif et al. (2024) \cite{munsifEfficientHybridClassification2024} & Heart Disease & GA-SVM-CNN & Achieved 98\% on UCI, 97\% on Z-Alizadeh Sani, and 86\% on CVD Dataset & UCI Heart Disease Dataset, Z-Alizadeh Sani Dataset, and CVD Dataset & - & -\\\hline
G. Chen et al. (2020) \cite{chenPredictionChronicKidney2020} & Chronic Kidney Disease & Adaptive Hybridized Deep Convolutional Neural Network (AHDCNN) on Pre-trained CNNs, autoencoders, and deep residual learning for feature extraction and model optimization & achieved high precision and recall, balanced F1-scores, and successful segmentation in 80\% of kidneys with a Dice score above 0.90. & Dataset obtained from \href{https://nihcc.app.box.com/v/DeepLesion} {link} & Enabled remote tracking of physical body state & - \\\hline
M. U. Nasir et al. (2022) \cite{nasirKidneyCancerPrediction2022} & Kidney Cancer  & IoMT-based TL technique with different DL Alex-net algorithms & achieved 99.8\% training accuracy and 99.20\% prediction accuracy. Validation accuracy was 93.75\%. & 3300 augmented samples across three grades (0, 1, 2) with 1100 samples per class & Online data collection from various hospitals & Data stored in secure private clouds on the blockchain \\\hline
K. Kumar et al. (2023) \cite{kumarDeepLearningApproach2023} & kidney disease & Hybrid fuzzy deep neural network (FDNN) with fuzzy logic principles &  achieved 99.23\% overall accuracy, compared to 97.46\% for the existing method &  Changhua Christian Hospital in Taichung, Taiwan, 5617 records & images captured by cameras to detect presence of disease  & - \\\hline
D. M. Alsekait et al. (2023) \cite{alsekaitComprehensiveChronicKidney2023} & Chronic Kidney Disease  & Stacking ensemble DL model combined outputs RNN, LSTM, and GRU in level 1, SVM in level 2 & RNN  achieved highest metrics: accuracy 97.88\%, precision 97.86\%, recall 97.95\%, F1-Score 97.88\% &  UCI, 400 cases, 24 features including 13 categorical and 11 numeric features & - & - \\\hline
T. Wang et al. (2018) \cite{wangPredictiveModelingProgression2018} & Alzheim. disease & employed a two-layer RNN model with 100 hidden units per layer &  accuracy of 0.9906 ± 0.0043, PPIA of 0.9894 ± 0.0074, and SPIA of 0.9912 ± 0.0039. &  5432 patients with probable AD, a subset of the NACC dataset & - & - \\\hline
Dua et al. (2020) \cite{duaCNNRNNLSTM2020} & Alzheimer & CNN, RNN, and LSTM with Bagging and weighted average ensemble & Accuracy: CNN, RNN, LSTM - 89.75\%, Ensemble - 92.22\% & OASIS-1 (416 MRI) and OASIS-2 (150 longitudinal MRI) & - \\ \hline
Chui et al. (2022) \cite{chuiMRIScansBasedAlzheimer2022} & Alzheimer & CNN, TL with domain knowledge, GAN for data augmentation & Accuracy increase: GAN 2.85–3.88\%, TL 2.43–2.66\%, Ablation study 1.8–40.1\% & OASIS-1, OASIS-2, OASIS-3 (416, 150, 1098 participants respectively) & - \\ \hline
Odusanmi et al. (2023) \cite{odusamiPixelLevelFusionApproach2023} & Alzheimer & Multimodal fusion with DWT, TL with VGG16, IDWT for final image & Accuracy: MRI - 81.25\% for AD/EMCI, AD/LMCI; PET - 93.75\% for AD/EMCI, AD/LMCI & sMRI and FDG-PET from ADNI, T1 volumes and 18F-FDG-PET with MMSE and CDR scores & - \\ \hline
Jungyoon Kim and Jihye Lim (2021) \cite{kimDeepNeuralNetworkBased2021} & dementia prediction & DNN with four hidden layers of 30 neurons each & AUC of 85.5\%,recall of 68.6\%, specificity of 82.1\%, accuracy of 81.9\%. & KNHANES consiting of 7031 participants  & - & - \\\hline
Y-A Choi et al. (2021) \cite{choiDeepLearningBasedStroke2021} & Stroke Disease  &  models like Bidirectional LSTM, CNN-LSTM, and CNN-Bidirectional LSTM. & The CNN-Bidirectional LSTM achioeved highest accuracy (94.0\%), precision (94.6\%), and F1-score (94.1\%) across various datasets &  EEG data collected from six channels & - & - \\\hline
Elbagoury et al. (2023) \cite{elbagouryHybridStackedCNN2023} & Stroke Disease  & GMDH-type polynomial network  &  accuracy of around 96.85\%. &  EMG Lower Limb Dataset, 24 patients performing three actions as normal or abnormal, and the mHealth Dataset & - & - \\\hline
C.T. Wu et Al. (2021) \cite{wuAcuteExacerbationChronic2021} & COPD & acute exacerbations of chronic obstructive pulmonary disorder (AECOPD) using various algorithms including DNN & DNN model outperforming others; F1 score of 0.9323, precision of 0.9393, specificity of 0.9253, sensitivity of 0.9452, AUROC of 0.9699, and accuracy of 0.9357 & clinical questionnaire data resulting in 5600 data points and 45 features were utilized & - & transmit data via the HTTPS protocol and encryption \\\hline
T. A. Khan et al. (2023) \cite{khanSecureIoMTDisease2023} & Lung Disease & applies TL using Google Net models in private edge clouds &achieves 98.8\% accuracy  &  LC25000 dataset, 25,000 images of cancer tissue in the lungs and colon &  datacollection using IoMT devices or sensors from hospitals &  Federated Learning (FL) approach for data privacy \\\hline
S. L. Jany Shabu et Al. (2023) \cite{shabuImprovedAdaptiveNeurofuzzy2023} & Lung cancer &Deep neural network, residual connections, ANFIS, IoMT, padding, pooling, dropout, convolution layers, 16 residual units, 2 fully connected layers &94.2\% accuracy in detection and prediction, mean absolute error: 52.3\% to 62.12\%   &  Private lung tumor dataset  &   a Fuzzy Expert System for Lung Cancer (FES-LC)  & - \\\hline
Hefu Xie et al. (2024) \cite{xieDeepLearningApproach2024} & Liver Disease & Fully connected CNN,, ALF, LR, KNN, SVM, DT, RF, XGBoost, GB & 94.8\% accuracy, superior generalization capabilities  &  Diverse patient indicators: age, gender, weight, obesity status, blood pressure, cholesterol, hepatitis status, family history  &  - & - \\\hline
Li et al. (2018) \cite{liEffectiveComputerAided2018} & Pancreatic & CAD model: (1)SLIC on CT pseudo-color images; (2) DT-PCA for feature extraction; (3) hybrid HFB-SVM-RF classifier & Acc.: 96.47\%, Sensibi: 95.23\%, Spec.: 97.51\%, evaluation on NIH dataset with mean Dice: 78.9\% and Jaccard: 65.4\% & PET/CT data from 80 patients, 1700 DICOM images per patient (CT: 512×512, PET: 128×128), labels confirmed by 3 experts  &  - & - \\ \hline
Asaturyan et al. (2019) \cite{asaturyanMorphologicalMultilevelGeometrical2019} & Pancreatic & MRI and CT scans using (1) enhance pancreas region; (2) 3D segmentation with max-flow and edge detection; (3) remove non-pancreatic contours & Mean DSC: 79.3\% ± 4.4\%. Evaluated on two MRI datasets with 216 and 132 volumes & Dataset with 82 CT volumes & - & -\\ \hline
Hu et al. (2021) \cite{huAutomaticPancreasSegmentation2021} & Pancreatic & DSD-ASPP-Net: Integrates DenseASPP with saliency-aware modules & Average DSC: 85.49\% ± 4.77\%, surpassing previous methods & Public NIH pancreas dataset and local hospital dataset & - \\ \hline
Anusha Ampavathi and T. Vijaya Saradhi (2021) \cite{ampavathiMultiDiseasepredictionFramework2021} & Diabetes, Hepatitis, Lung Cancer, Liver Tumor, Alzheimer's, Parkinson's, Heart Disease &  Hybrid JA-MVO-RNN + DBN algorithm compared with SVM, KNN, NN, RNN, DBN, and RNN + DBN & 
Diabetes (Prec.: 0.97753, F1-Score: 0.95604, MCC: 0.93528), Hepatitis (Prec.: 0.98684, F1-Score: 0.88757, MCC: 0.84824), Liver Tumor (Prec.: 0.97753, F1-Score: 0.95604, MCC: 0.93528), Alzheimer's (Prec.: 1, F1-Score: 0.96089, MCC: 0.94403) & UCI: Hepatitis (19 attributes), Lung Cancer (56 attributes), Liver Tumor (7 attributes), and Alzheimer's (34 attributes) & - & - \\\hline
R. Daid et al. (2021) \cite{daidEffectiveMechanismEarly2021} & Heart Disease, Stroke, Lung Cancer etc & DL-based multi-layer CNN &  accuracy values of 0.973 (80\% train, 20\% test), 0.949 (70\% train, 30\% test), and 0.969 (85\% train, 15\% test) &  illness severity (measured by the Harvey Bradshaw Index) from 600,000 patients & -&- \\\hline
A. S. Prakaash et al. (2022) \cite{prakaashDesignDevelopmentModified2022} & Alzheimer’s disease, cancers etc. &  ensemble learning like DNN, SVM, and RNN with weighted RBM features. Optimizes hidden neurons in DNN using the ASR-CHIO algorithm & F1-scores (0.915 to 0.931), MCC (0.760 to 0.874), specificity (0.912 to 0.938), NPV (0.901 to 0.941), and high accuracy (0.913 to 0.938) & COVID-19- github; (EEG Eye State andDiabetic)- UCI (Epileptic Seizure  and Stroke Prediction)- kaggle; Heart- datahub & -&-  \\\hline
S. Thandapani et al. (2023) \cite{thandapaniIoMTDeepCNN2023} & Pandemic Support System &  various DCNN such as ResNet 50, ResNet 100, ResNet 101, VGG 16, and VGG 19 & ResNet 101 outperformed others: CT images (Precision 97\%, Recall 92\%, F1-Score 96\%, Accuracy 97\%) for X-ray images (Precision 98\%, Recall 92\%, F1-Score 96\%, Accuracy 98\%) & from Kaggle, GitHub, UCI, DBT, hospitals, and scan centers, with dataset sizes of 1700 CT images and 2200 X-ray images.synthetic data included 7455 CT and 8900 X-ray & Patient temp and pulse rate sensors via internet and cloud & - \\\hline
N. Nigar et al. (2023) \cite{nigarIoMTMeetsMachine2023} & COVID-19, pneumonia, diabetes, heart disease, brain tumor, and Alzheimer's &  five pre-trained deep CNN models - VGG16, VGG19, ResNet, DenseNet, and Inception-v3 trainned using HPoptimization with Bayesian optimization and XGBoost algorithm &  COVID-19: VGG-16 acc. 80\%, prec. 80\%, recall 80\%, F1 score  80\%;pneumonia VGG-16 model reached an accuracy of 87.18\%, recall of 96\%, F1 score of 90\%;heart: VGG-19 val. acc. 88.46\%  & : COVID-19, diabetes, heart diseases (real-time); brain tumor, pneumonia (Kaggle, real data); Alzheimer's (OASIS, real datasets) & medical devices and sensors integration & Elliptic Curve Cryptography (ECC) for authentication and encryption of data \\\hline
M. K. Chowdary et al. (2023) \cite{chowdaryMultipleDiseasePrediction2023} & Heart disease, breast cancer, kidney etc & Random ForestRF,DT, GB,LR, XGBoost, SVM, KNN, and DL techniques like VGG19 CNN  &  VGG19 achieve highest acc. 98.5\%. & Kaggle: heart disease, diabetes, UCI ML resipository: breast cancer, CKD from the National Institute of Health (NIH), and pneumonia from the Guangzhou Women and Children's Medical Center & -& -  \\\hline
\end{longtable}

\bibliographystyle{elsarticle-num} 

% bibliography database
\bibliography{mysurvey}

%% Authors Biography

\begin{biography}
\noindent
\begin{tabularx}{\textwidth}{X X}
\bio{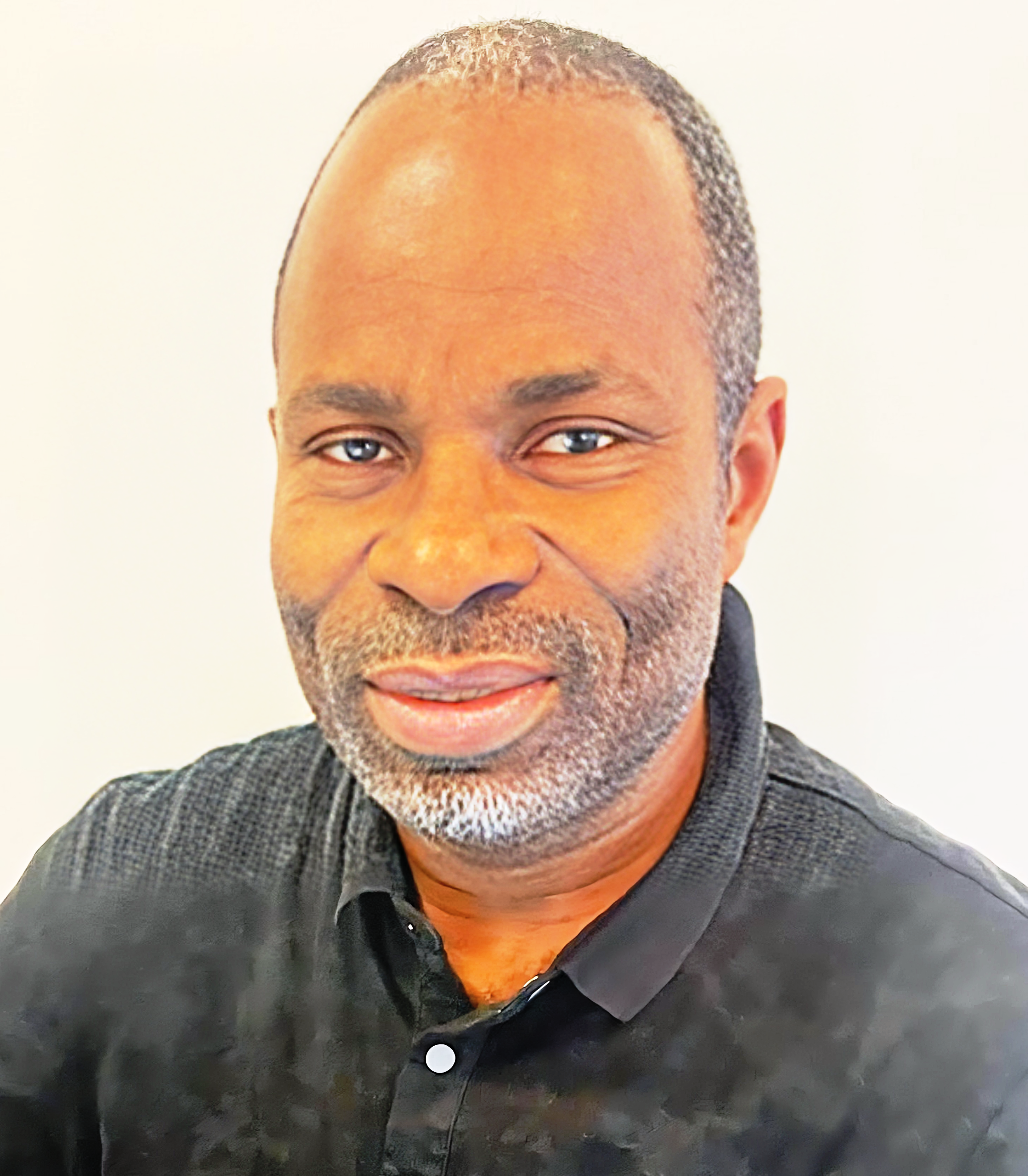}
Akeem Temitope Otapo earned both his B.Sc. and M.Sc. degrees in electronic and computer engineering from Lagos State University, Nigeria, in 2004 and 2016, respectively. He currently serves as a Senior Lecturer in the Department of Computer Engineering at Yaba College of Technology, Nigeria. Additionally, he is pursuing his doctoral degree at the Laboratoire Images, Signaux et Systèmes Intelligents (LISSI) at Université Paris-Est Créteil, France. His research interests include the Internet of Things (IoT) and Embedded Systems, AI models, and Biomedical Engineering. Mr. Otapo is a registered engineer with the Council for the Regulation of Engineering in Nigeria (COREN) and a member of the International Association of Engineers (IAENG), the Nigeria Computer Society (NCS), and the Nigeria Institution of Management (NIM).
\endbio
&
\bio{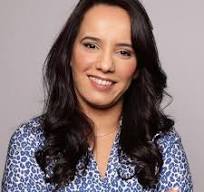}
Alice Othmani is a Maître de Conférences (Associate Professor) at the University of Paris-Est Créteil since 2017. She holds a Ph.D. in Computer Vision from the University of Burgundy, awarded in 2014. She has worked at various institutions, including École Normale Supérieure Ulm Paris, the University of Medicine of Auvergne, and the Agency for Science, Technology and Research (A*STAR) in Singapore. Her expertise includes image and signal processing, pattern recognition, computer vision, and machine learning. She is interested in the development of intelligent systems for healthcare.
\endbio
\\[2ex]
\bio{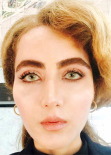}
Ghazaleh Khodabandelou is an associate professor at the University of Paris-Est Créteil. She received her Ph.D. degree from Paris Sorbonne University. Since 2022, she has been collaborating as a researcher with the University of Washington. Her research interests include Optimization and Mathematical Modeling to develop Artificial Intelligence solutions in Ambient Assisted Living, eHealth, Robotics, and Genomics.
\endbio
&
\bio{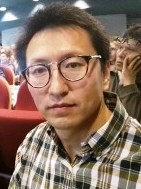}
Zuheng Ming received his B.Sc. degree from Hunan University and his MSc degree from Beijing Institute of Technology. He obtained his Ph.D. from Gipsa-lab at the University of Grenoble Alpes. He is currently a tenured assistant professor (Maître de conférence) of computer science at the Laboratoire L2TI, Institut Galilée, Université Sorbonne Paris Nord (formerly Université Paris-XIII). Before joining Université Sorbonne Paris Nord, Dr. Ming was a lecturer-researcher (Enseignant-Chercheur Contactuel) in the computer science department at L3i, La Rochelle University. He also completed postdoctoral positions at L3i, La Rochelle University. Dr. Ming’s research interests include computer vision, multimodal learning, deep learning, document analysis, biometric systems, remote sensing, and medical image processing.
\endbio
\end{tabularx}

\end{biography}

\end{document}